\documentclass{article}

\usepackage{arydshln}
\usepackage{microtype}
\usepackage{graphicx}
\usepackage{subcaption}
\usepackage{booktabs} 
\usepackage{multirow}
\usepackage{hyperref}

\usepackage[normalem]{ulem}
\useunder{ine}{}{}
\useunder{\uline}{\ul}{}

 \usepackage[preprint]{icml2026}

\usepackage{amsmath}
\usepackage{amssymb}
\usepackage{mathtools}
\usepackage{amsthm}

\usepackage[capitalize,noabbrev]{cleveref}

\theoremstyle{plain}

\theoremstyle{definition}

\theoremstyle{remark}

\usepackage[textsize=tiny]{todonotes}

\icmltitlerunning{Trifuse: Enhancing Attention-Based GUI Grounding via Multimodal Fusion}

\begin{document}

\twocolumn[
\icmltitle{Trifuse: Enhancing Attention-Based GUI Grounding via Multimodal Fusion}

\begin{icmlauthorlist}
	
	\icmlauthor{Longhui Ma}{nudt}
	\icmlauthor{Di Zhao}{nudt}
	\icmlauthor{Siwei Wang}{ams}
	\icmlauthor{Zhao Lv}{ams}
	\icmlauthor{Miao Wang}{ams}
	
\end{icmlauthorlist}

\icmlaffiliation{nudt}{
	College of Computer Science and Technology,
	National University of Defense Technology,
	Changsha, China
}

\icmlaffiliation{ams}{
	Intelligent Game and Decision Lab,
	Academy of Military Sciences,
	Beijing, China
}

\icmlcorrespondingauthor{Siwei Wang}{wangsiwei13@nudt.edu.cn}

\icmlkeywords{GUI Agents, Multimodal Fusion, Attention-based Grounding, Human-Computer Interaction}

\vskip 0.3in
]



\printAffiliationsAndNotice{}  


%
\begin{abstract}
GUI grounding maps natural language instructions to the correct interface elements, serving as the perception foundation for GUI agents.
Existing approaches predominantly rely on fine-tuning multimodal large language models (MLLMs) using large-scale GUI datasets to predict target element coordinates, which is data-intensive and generalizes poorly to unseen interfaces. 
Recent attention-based alternatives exploit localization signals in MLLMs attention mechanisms without task-specific fine-tuning, but suffer from low reliability due to the lack of explicit and complementary spatial anchors in GUI images. To address this limitation, we propose Trifuse, an attention-based grounding framework that explicitly integrates complementary spatial anchors. Trifuse integrates attention, OCR-derived textual cues, and icon-level caption semantics via a Consensus-SinglePeak (CS) fusion strategy that enforces cross-modal agreement while retaining sharp localization peaks. Extensive evaluations on four grounding benchmarks demonstrate that Trifuse achieves strong performance without task-specific fine-tuning, substantially reducing the reliance on expensive annotated data. Moreover, ablation studies reveal that incorporating OCR and caption cues consistently improves attention-based grounding performance across different backbones, highlighting its effectiveness as a general framework for GUI grounding.
\end{abstract}

\begin{figure}
	\centering
	\includegraphics[width=0.9\linewidth]{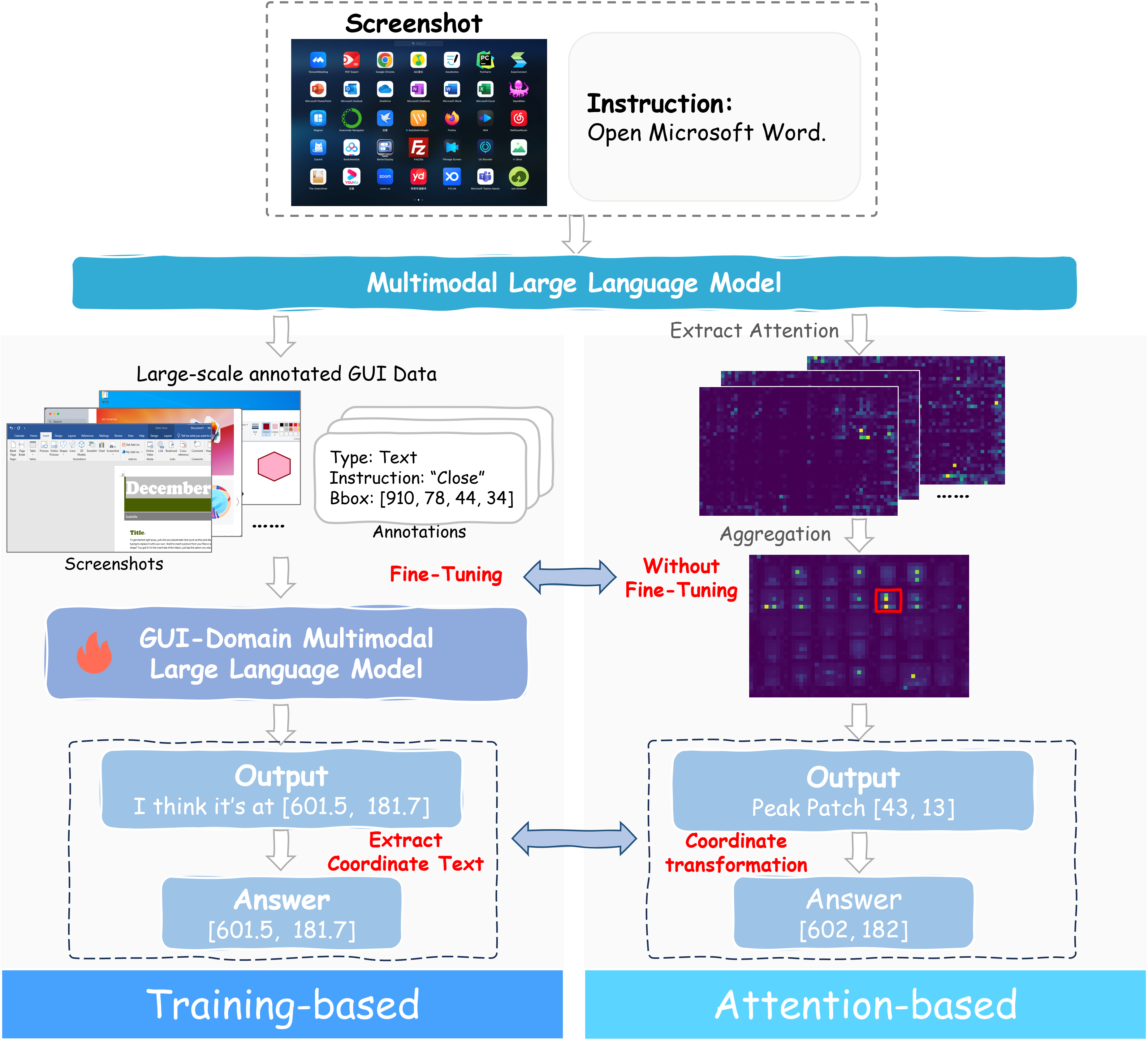}
	\caption{Comparison between training-based and attention-based methods for GUI grounding.}
	
	\label{fig:intro}
\end{figure}

\section{Introduction}

Graphical User Interfaces (GUIs) represent an essential component of contemporary digital systems, enabling users to interact visually with software through GUI elements~\cite{shneiderman2010designing,johnson2020designing}.
Recent advances in multimodal large language models (MLLMs)~\cite{gpt4o,gemini, qwen2.5vl} have driven increased research interest toward the development of intelligent GUI agents capable of understanding and interacting within GUI environments~\cite{guisurvey,guisurvey2}.
These agents offer significant potential for automating software tasks, supporting accessibility
requirements, and enabling generalization across diverse platforms, thereby opening new avenues for advancements in human-computer interaction~\cite{showui,uground}.

Accurately grounding natural language instructions to corresponding GUI elements remains a critical requirement for GUI agents ~\cite{tang2025survey}.
Recent work predominantly employs training-based methods, where MLLMs are fine-tuned on large-scale GUI datasets to directly predict target element coordinates. 
While these methods~\cite{showui, uitars} achieve strong grounding benchmark performance~\cite{seeclick, osworldg}, they require substantial annotated data and often exhibit limited generalization. 
Adapting to new applications, layouts, or operating systems typically necessitates additional data collection and retraining~\cite{transbench}.
Attention-based methods offer an alternative by leveraging the internal attention mechanisms of MLLMs for localization. 
Prior work demonstrates that MLLMs' attention naturally focuses on task-relevant visual regions in natural images~\cite{mllmknow, attentionhead}.
TAG~\cite{tag} first applies this observation to GUI environments by selecting informative tokens and attention heads to construct attention maps for grounding without task-specific fine-tuning. 
While demonstrating the feasibility of attention-based grounding,
TAG still lags behind supervised models in grounding performance.
Subsequent attention-based methods, including GUI-Actor and GUI-AIMA~\cite{guiactor, guiaima}, introduce learnable tokens to guide attention, improving accuracy but reintroducing training requirements and data dependencies.
Training-free approaches like TAG remain less accurate than training-based methods.

This naturally raises a key question: \textit{whether accurate GUI grounding can be achieved using attention mechanisms without task-specific fine-tuning}.
We believe that the failure of attention-based GUI grounding is not due to a lack of semantic understanding, as TAG~\cite{tag} has achieved certain performance, but rather to a lack of complementary spatial cues.
Recent studies~\cite{omniparser, mp-gui, dimogui} show that incorporating additional GUI information—such as text, visual semantics, and structural relations—substantially improves grounding performance, motivating a multimodal fusion approach.
Building on these insights, we propose Trifuse, an attention-based GUI grounding framework that integrates three complementary modalities: MLLM attention, OCR-based textual cues and icon caption-based visual semantics.
For each modality, Trifuse uses dedicated mechanisms to extract corresponding heatmaps. 
Then, Trifuse fuses these modality heatmaps through a Consensus-SinglePeak (CS) strategy that emphasizes consensus signals across modalities while suppressing noise.
Compared with other attention-based methods, 
experiments show that Trifuse outperforms prior training-free methods~\cite{tag, omniparser} on four grounding benchmarks~\cite{seeclick, screenspotpro, osworldg}. 
Furthermore, when integrated with training-based models~\cite{guiactor, guiaima}, Trifuse further improves localization accuracy, demonstrating the complementary and generalization value of multimodal fusion even for fine-tuned MLLMs.
Our contributions are:
\begin{itemize}
	\item We propose Trifuse, an attention-based GUI grounding framework that fuses three modalities: MLLM attention, OCR-based textual cues and icon caption-based visual semantics to improve grounding accuracy without requiring GUI-specific annotated data.
	
	\item We introduce an efficient attention extraction strategy that selects informative tokens and attention heads from MLLMs, enabling more precise and robust
	grounding.
	
	\item We design a Consensus-SinglePeak (CS) fusion strategy that
	jointly models cross-modal agreement and modality-specific discriminative
	signals, effectively suppressing noise while preserving informative peaks.
	Extensive experiments across multiple benchmarks validate the effectiveness
	and generality of the proposed fusion strategy.
\end{itemize}

\section{Related Work}
\subsection{GUI Agents}
Recent advances in large language models (LLMs) and vision-language integration have enabled powerful MLLMs such as GPT-4o~\cite{gpt4o}, Gemini~\cite{gemini}, and open-source alternatives like Qwen-VL~\cite{qwen2vl, qwen2.5vl}, which demonstrate strong capabilities across vision-language tasks. 
These MLLMs provide a foundation for GUI agents that understand user instructions and interact with GUIs to complete tasks.
Early GUI agents primarily operated in web and mobile platforms with access to structured representations such as HTML DOM trees or accessibility hierarchies~\cite{mind2web, wang2024mobile}, enabling precise element identification through explicit structural information. 
Recent work has shifted toward vision-centric paradigms that rely solely on visual observations~\cite{showui, uground, aguvis}, eliminating dependence on platform-specific APIs or structured data.
\subsection{Grounding in GUI Agents}
GUI grounding aims to map natural language instructions to target GUI elements in screenshots.
Training-based methods are the most commonly used approaches, which fine-tune MLLMs on large GUI datasets to predict element coordinates~\cite{aguvis, seeclick, osatlas}.
While achieving strong benchmark performance~\cite{uitars}, these methods require substantial annotated data to align spatial coordinates with visual elements and often generalize poorly without additional retraining.
Reinforcement learning approaches~\cite{gui_rl} attempt to reduce data requirements through GUI-specific reward signals but remain computationally intensive.
Training-free methods attempt to locate target elements without task-specific fine-tuning~\cite{omniparser, dimogui,som}.
Among them, attention-based approaches exploit the internal attention mechanisms
of MLLMs, which have been shown to naturally focus on task-relevant visual
regions~\cite{mllmknow, attentionhead}.
~\cite{tag} first exploit this observation in GUI environments, demonstrating that MLLMs attention exhibits GUI grounding capability.
However, performance is limited by heuristic token and head selection strategies.
To improve accuracy, subsequent attention-based methods introduce specialized learnable tokens ~\cite{guiactor, guiaima}, but require task-specific fine-tuning.

\begin{figure*}
	\centering
	\includegraphics[width=0.85\linewidth]{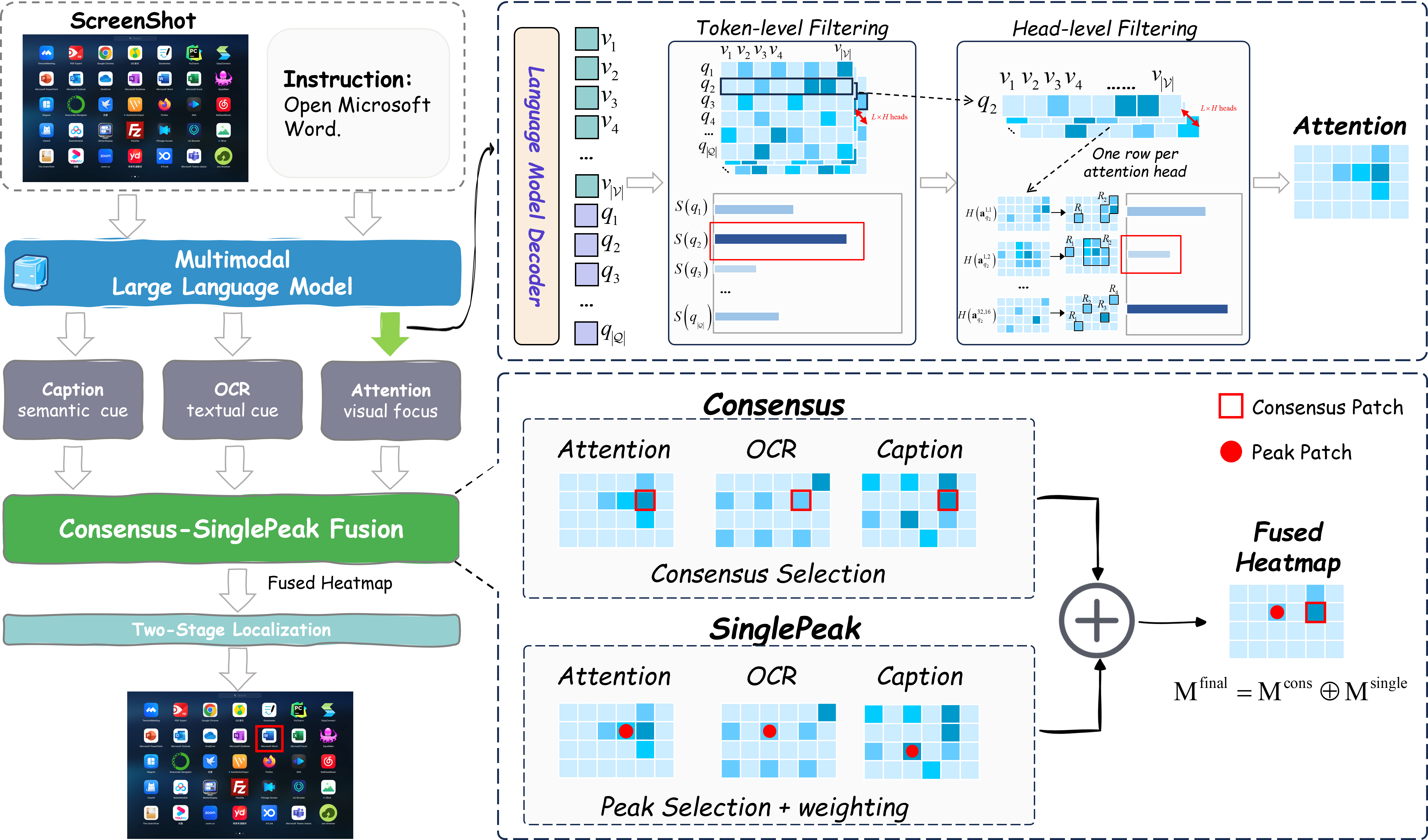}
	\caption{Overview of our Trifuse framework.
		Trifuse consists of three main components:
		(1) a modality extraction module that derives complementary grounding cues, including attention-based signals from
		MLLMs, textual cues from OCR, and icon-level visual semantics from captioning;
		(2) a Consensus-SinglePeak (CS) fusion module that integrates these
		modality-specific heatmaps by jointly modeling cross-modal agreement and
		modality-specific discriminative peaks; and
		(3) a two-stage localization module that progressively refines the fused
		grounding map through cropping and zoom-in operations to accurately identify
		the target GUI element.}
	\label{fig:2framework}
\end{figure*}

\section{Method}
\label{sec:method}
Trifuse mainly consists of three parts: modality extraction, modality fusion, and localization. 
The modality extraction stage constructs three complementary grounding heatmaps:
(1) attention-based heatmap via fine-grained token and head filtering, (2) OCR-based heatmap from detected textual elements, and (3) caption-based heatmap from icon-level visual semantics. 
The modality fusion stage integrates these modalities through the Consensus-SinglePeak (CS) strategy.
Finally, a two-stage localization strategy is applied to the fused heatmap to locate the target element.

\label{sec:filtering}
\subsection{Modality Extraction}

\subsubsection{Attention Modality}
MLLMs produce multi-head attention maps across layers.
However, directly leveraging all attention information for grounding performs poorly due to noise from irrelevant tokens and weakly-aligned heads~\cite{tag}. 
We address this through a two-level filtering scheme: token-level filtering identifies task-relevant query tokens, and head-level filtering identifies spatially informative attention heads.
The filtered attention signals are subsequently aggregated to construct a reliable attention-based grounding heatmap.

\paragraph{Preliminaries}
We consider an MLLM with $L$ layers and $H$ attention heads that processes GUI screenshots $V$ and user instructions $Q$. $V$ is represented as a sequence of visual patch tokens $\mathcal{V} = [v_1, \dots, v_{|\mathcal{V}|}]$ after the vision encoder. The user instructions are tokenized as $\mathcal{Q} = [q_1, \dots, q_{|\mathcal{Q}|}]$. 
At layer $l$ and head $h$, the model computes a self-attention matrix $\mathbf{A}^{l,h}$. For each query token $q_i$, we extract its patch-wise attention vector $\mathbf{a}^{l,h}_{q_i} \in \mathbb{R}^{|\mathcal{V}|}$, where the $j$-th element of $\mathbf{a}^{l,h}_{q_i}$ indicates how strongly $q_i$ attends to visual patch $v_j$.

\paragraph{Token-level Filtering}
User instructions typically consist of multiple tokens with varying relevance to the target elements.
While certain tokens explicitly describe the target object or its attributes, others are only implicitly related, and function words such as prepositions or articles often provide little grounding signal.
Prior studies on referring expression comprehension and vision–language grounding have shown that treating all tokens equally introduces substantial noise and degrades localization performance~\cite{yu2018mattnet,qiao2020referring}.
This motivates the need for selectively emphasizing informative tokens during grounding.

To identify the most relevant tokens, we measure the alignment between each text token $q_i$ and the visual content by computing a token-image relevance score:
\begin{equation}
	S(q_i) = \sum_{v_j \in \mathcal{V}} \mathbb{I}\!\left[\cos(q_i, v_j) > \tau_v \right] \cdot \cos(q_i, v_j),
\end{equation}
where $q_i$ and $v_j$ denote the embedding of token $q_i$ and visual patch $v_j$, respectively. A higher score indicates stronger correlation with image regions, suggesting greater relevance for grounding. 
$\tau_v$ reduces interference from irrelevant patches.
Then, we retain the top-$k$ tokens with the highest relevance scores,
forming the filtered token set $\mathcal{Q}_{\text{filter}}$.

\paragraph{Head-level Filtering}
Recent work has shown that not all attention heads are equally suitable for visual grounding in natural images~\cite{xiao2024towards, attentionhead}, and similar observations hold in the GUI domain~\cite{tag}. Only a subset of heads exhibits spatially concentrated attention patterns useful for grounding, while others distribute attention diffusely across the image (see Appendix~\ref{lab:attention} for cases).

To identify grounding-relevant attention heads, we use spatial entropy to quantify how concentrated each head’s attention distribution is over spatial regions.
Specifically, we group the attention heatmap into a set of spatially connected
regions using connected component analysis implemented via a union--find algorithm, denoted as
$\mathcal{R} = \{ r_1, r_2, \dots, r_{|\mathcal{R}|} \}$.
For a given attention head $(l,h)$ and a selected token
$q_i \in \mathcal{Q}_{\text{filter}}$, the spatial entropy of its attention
distribution $\mathbf{a}^{l,h}_{q_i}$ is defined as:
\begin{equation}
	H\!\left(\mathbf{a}^{l,h}_{q_i}\right)
	= - \sum_{r_j \in \mathcal{R}} R_j \log R_j
\end{equation}
where $R_j$ represents the normalized attention mass assigned to the $j$-th
connected region:
\begin{equation}
	R_j =
	\frac{\sum_{v_k \in r_j} \left(\mathbf{a}^{l,h}_{q_i}\right)_k}
	{\sum_{r_m \in \mathcal{R}} \sum_{v_k \in r_m}
		\left(\mathbf{a}^{l,h}_{q_i}\right)_k}
\end{equation}
Lower entropy indicates that the attention is concentrated on fewer connected
regions, suggesting that the head is more likely to focus on the target GUI elements. 
We retain the top-$k$ attention heads with the lowest spatial entropy,
forming the grounding-relevant head set $\mathcal{H}_{\text{filter}}$.
Head selection is performed independently for each token in $\mathcal{Q}_{\text{filter}}$.
\paragraph{Weighted Aggregation}
After token and head filtering, we construct the final attention heatmap
$\mathbf{a}^{\text{attn}}$ via weighted aggregation over $\mathcal{Q}_{\text{filter}}$ and $\mathcal{H}_{\text{filter}}$.
Specifically, token weights are computed by applying a softmax over
token-image relevance scores, while head weights are obtained by applying
a softmax over the negative spatial entropy.

\begin{equation}
	\scalebox{0.7}{$
		\begin{aligned}
			\mathbf{a}^{\text{attn}}
			=
			\sum_{q_i \in \mathcal{Q}_{\text{filter}}}
			\mathrm{softmax}\!\left(S(q_i)\right)
			\sum_{(l,h) \in \mathcal{H}_{\text{filter}}(q_i)}
			\mathrm{softmax}\!\left(-H(\mathbf{a}^{l,h}_{q_i})\right)
			\mathbf{a}^{l,h}_{q_i}
		\end{aligned}
		$}
\end{equation}

\subsubsection{OCR and Caption Modality}

\paragraph{OCR Modality.}
GUI images often contain rich textual information, which plays a critical role
in text-related grounding tasks~\cite{textvqa}. Accurately extracting and
aligning textual content from GUI screenshots is therefore essential for precise grounding.

We employ an off-the-shelf OCR engine to extract textual elements from a given
GUI image $V$. The OCR engine outputs a set of detected text instances
$
\mathcal{T} = \{ t_1, t_2, \dots, t_{|\mathcal{T}|}\},
$
where each instance $t_k = \{ b_k, s_k, c_k \}$ consists of a bounding box
$b_k$, a recognized text string $s_k$, and a detection confidence score
$c_k \in [0,1]$.
To align OCR results with the user instructions, we compute the semantic
relevance between each recognized text string and the filtered query tokens.
Specifically, the OCR relevance score for instance $t_k$ is defined as:
\begin{equation}
	r_k^{\text{ocr}} =
	\cos\!\big( \mathbf{e}(s_k), \mathcal{Q}_{\text{filter}} \big)
	\cdot c_k,
\end{equation}
where $\mathbf{e}(\cdot)$ denotes the text embedding function.
We then project the OCR relevance scores onto the visual patch grid.
For each visual patch $v_j$, the OCR-based heatmap value aggregates contributions
from all OCR bounding boxes that overlap with $v_j$:
\begin{equation}
	a^{\text{ocr}}_j
	=
	\sum_{k:\, v_j \cap b_k \neq \emptyset} r_k^{\text{ocr}}.
\end{equation}
After normalization, the resulting vector
$\mathbf{a}^{\text{ocr}} \in \mathbb{R}^{|\mathcal{V}|}$
represents the OCR-based heatmap.

\paragraph{Caption Modality.}
GUI screenshots contain abundant non-textual visual elements, such as buttons and graphical controls, which are essential for grounding instructions
that do not explicitly mention text~\cite{aitw, rico}.
To capture such visual semantics beyond textual cues, we employ a pretrained
icon detection model to localize GUI elements and generate semantic descriptions.

Given an input image $V$, the icon detection model outputs a set of detected icon
instances
$
\mathcal{C} = \{ c_1, c_2, \dots, c_{|\mathcal{C}|} \},
$
where each instance $c_k = \{ b_k, d_k, c_k^{\text{conf}} \}$ consists of a
bounding box $b_k$, a generated caption or semantic description $d_k$, and a
detection confidence score $c_k \in [0,1]$.
To align icon semantics with the user instructions, we compute the semantic
similarity between each generated caption and the filtered query tokens.
The caption relevance score for instance $c_k$ is defined as:
\begin{equation}
	r_k^{\text{cap}} =
	\cos\!\big( \mathbf{e}(d_k), \mathcal{Q}_{\text{filter}} \big)
	\cdot c_k.
\end{equation}
Similar to the OCR modality, we project the caption relevance scores onto the
visual patch grid. For each visual patch $v_j$, the caption-based heatmap value
aggregates contributions from all detected icon regions that overlap with
$v_j$:
\begin{equation}
	a^{\text{cap}}_j
	=
	\sum_{k:\, v_j \cap b_k \neq \emptyset} r_k^{\text{cap}}.
\end{equation}
After normalization, the resulting vector
$\mathbf{a}^{\text{cap}} \in \mathbb{R}^{|\mathcal{V}|}$
represents the patch-wise caption grounding heatmap.

%
%

\subsection{Modality Fusion}
After obtaining the modality-specific heatmaps
$\mathbf{a}^{\text{attn}}$, $\mathbf{a}^{\text{ocr}}$, and
$\mathbf{a}^{\text{cap}}$, an effective fusion strategy is required to
highlight the target region while avoiding bias toward any single modality.
Simple averaging is vulnerable to noisy but dominant modalities, whereas pure consensus-based fusion may suppress
informative modality-specific cues.
These limitations motivate a fusion strategy that explicitly separates
cross-modal agreement from discriminative responses.
Therefore, we propose a Consensus-SinglePeak (CS) fusion strategy that integrates two complementary types of signals:
cross-modal consensus, which emphasizes regions consistently supported
by multiple modalities, and modality-specific peaks, which preserve
strong and discriminative responses unique to individual modalities.
By jointly modeling these two aspects, CS fusion achieves a balanced trade-off
between robustness and sensitivity.

Formally, the final fused heatmap is computed as:
\begin{equation}
	\mathbf{M}^{\text{final}} = \mathbf{M}^{\text{cons}} \oplus \mathbf{M}^{\text{single}},
\end{equation}
where $\mathbf{M}^{\text{cons}} \in \mathbb{R}^{|\mathcal{V}|}$ captures the cross-modal consensus and $\mathbf{M}^{\text{single}} \in \mathbb{R}^{|\mathcal{V}|}$ preserves modality-specific peaks. The operator $\oplus$ denotes an element-wise
fusion operation that combines the two signals by element-wise averaging.

\paragraph{Consensus}
The consensus term identifies spatial regions where multiple modalities exhibit strong agreement. We compute the consensus heatmap by aggregating modality-specific heatmaps
through element-wise multiplication:
\begin{equation}
	M_j^{\text{cons}} = a_j^{\text{attn}} \odot a_j^{\text{ocr}} \odot a_j^{\text{cap}},
\end{equation}
where $a_j^{\text{attn}}$, $a_j^{\text{ocr}}$, and $a_j^{\text{cap}}$ denote heatmap values from attention, OCR, and caption modalities at patch position $j$, respectively. This multiplicative formulation amplifies regions that are consistently supported by all modalities.

\paragraph{Single-Peak}
While the consensus term emphasizes regions jointly supported by multiple
modalities, modality-specific peaks often provide crucial discriminative cues
that may be absent from other modalities.
For example, OCR can uniquely localize a target text element that is weakly
attended by the attention or caption modalities.
To preserve such informative signals while suppressing spurious responses, we
introduce a confidence-aware single-peak selection mechanism.

For each modality $s \in \{\text{attn}, \text{ocr}, \text{cap}\}$ and patch
position $j$, we first identify candidate peaks whose heatmap responses exceed
a modality-specific threshold, i.e.,
$a_j^s > \tau_s$.
Let $P_s = \{ j \mid a_j^s > \tau_s \}$ denote the peak set for modality $s$.
To estimate the reliability of each peak, we measure its relative support from
the remaining modalities:
\begin{equation}
	\text{conf}_{s,j}
	=
	\sigma\!\left(
	\alpha \cdot
	\frac{\sum_{s' \neq s} a_j^{s'}}{a_j^s + \varepsilon}
	-
	\beta
	\right),
\end{equation}
where $\sigma(\cdot)$ denotes the sigmoid function, $\varepsilon$ is a small
constant for numerical stability, and $\alpha, \beta$ control the sensitivity
of the confidence estimation.
Intuitively, the confidence score is high when a modality-specific peak is
supported by non-negligible responses from other modalities, and low when the
peak is isolated and potentially noisy.

Based on the confidence score, we assign a dynamic weight to each peak:
\begin{equation}
	W_{s,j} = 1 + \lambda \big( 2\,\text{conf}_{s,j} - 1 \big),
\end{equation}
where $\lambda \geq 0$ controls the strength of peak amplification.
High-confidence peaks are amplified ($W_{s,j} > 1$), whereas low-confidence
peaks are softly down-weighted ($W_{s,j} < 1$).
The single-peak heatmap aggregates all weighted modality-specific peaks:
\begin{equation}
	M_j^{\text{single}}
	=
	\sum_{s \in \{\text{attn}, \text{ocr}, \text{cap}\}}
	\mathbb{I}(j \in P_s)\,
	W_{s,j}\,
	a_j^s .
\end{equation}

Together with the consensus term, the Single-Peak component enables CS fusion
to retain highly discriminative modality-specific cues while maintaining
robustness to noise.
After obtaining the fused heatmap $\mathbf{M}^{\text{final}}$, we next describe
how to localize the target GUI elements from it.

\subsection{Two-Stage Localization}
Directly selecting the peak from the fused heatmap
$\mathbf{M}^{\text{final}}$ often yields suboptimal localization accuracy
(Section~\ref{label:ablation}), primarily due to the limited spatial granularity
of visual tokens in MLLMs.
High-resolution GUI screenshots are typically downsampled to meet GPU memory
constraints, resulting in coarse patch tokenization that degrades fine-grained
grounding.

To address this issue, we adopt a simple yet effective two-stage zoom-in
localization strategy~\cite{screenspotpro, appagent} without requiring additional training.
In the first stage, we perform inference on the downsampled full screenshot to
obtain the fused heatmap $\mathbf{M}^{\text{final}}$, and identify a coarse
target region $R_{\text{coarse}}$ by selecting the patch with the maximum
response.
In the second stage, we crop a window centered at $R_{\text{coarse}}$ from the
original high-resolution image, resize it to the model’s input resolution, and
re-run inference to obtain a refined prediction.
This hierarchical strategy decouples coarse region identification from
fine-grained localization, substantially improving precision on
high-resolution displays without architectural modifications or task-specific retraining.


\section{Experiments}

\paragraph{Implementation Details.}
We employ Qwen2.5-VL-3B-Instruct~\cite{qwen2.5vl} as the MLLM backbone for attention extraction.
The model consists of 32 transformer layers, each with 16 attention heads.
Following the token and head filtering strategy described in
Section~\ref{sec:filtering}, we select the top-1 semantically relevant token and aggregate attention from the top-6 spatially informative heads.
For the OCR and caption modality, we adopt PaddleOCR v4~\cite{paddleocr} for
text detection and recognition, and OmniParser~\cite{omniparser} for icon
localization and semantic caption generation, respectively.
To compute semantic similarity between textual content (OCR results or icon
captions) and the user instructions, we use the BGE-M3~\cite{bgem3} embedding
model.
For two-stage localization, we perform a single zoom-in iteration.
Specifically, a cropped window of size $\frac{W}{2} \times \frac{H}{2}$ is
extracted from the original image and centered at the coarse prediction, where
$W \times H$ denotes the input image resolution.
The cropped region is then resized to the model’s input resolution for
refined inference.
We follow standard practice and define training-free as no task-specific fine-tuning on GUI grounding datasets. All auxiliary models (OCR, icon parser, embedding models) are used off-the-shelf without adaptation.

\paragraph{Evaluation Benchmarks}
We evaluate Trifuse on four established GUI grounding benchmarks: ScreenSpot, ScreenSpot-v2~\cite{seeclick}, ScreenSpot-Pro~\cite{screenspotpro}, and OSWorld-G~\cite{osworldg}. These benchmarks collectively assess grounding performance across diverse platforms (mobile, desktop, web), resolutions, and task complexities. We report Element Accuracy as the primary metric, which measures the proportion of predictions where the predicted click point falls within the ground-truth bounding box of the target elements. Additional benchmark details are provided in Appendix~\ref{label:benchmarks}.

\paragraph{Baselines}
We compare Trifuse against four categories of GUI grounding methods representing distinct paradigms. \textbf{General models} are pretrained vision-language models without GUI-specific fine-tuning, including GPT-4o~\cite{gpt4o} and Qwen2.5-VL~\cite{qwen2.5vl}. \textbf{Supervised fine-tuning (SFT) models} are VLMs fine-tuned on GUI-annotated datasets, including UGround~\cite{uground}, Aguvis~\cite{aguvis}, ShowUI~\cite{showui}, UI-TARS~\cite{uitars}, and JEDI~\cite{osworldg}. \textbf{Reinforcement learning (RL) models} employ policy gradient methods with GUI-specific reward signals, including UI-R1~\cite{uir1}, GUI-G1~\cite{guig1}, and GUI-G$^2$~\cite{guig2}. 
\textbf{Attention-based methods} exploit attention mechanisms of MLLMs to
localize target GUI elements. 
This category includes the training-free method
TAG~\cite{tag}, as well as training-based approaches including GUI-Actor~\cite{guiactor} and GUI-AIMA~\cite{guiaima}.

Among these baselines, attention-based methods constitute our primary
comparison group, as Trifuse similarly leverages attention for
grounding while introducing multimodal fusion to enhance localization accuracy.
To isolate the contribution of the proposed fusion strategy from backbone
capacity, we evaluate Trifuse with multiple backbone MLLMs.
Specifically, for comparisons with training-based methods, we replace the
Qwen2.5-VL-3B-Instruct backbone with the corresponding fine-tuned
GUI-Actor-3B, GUI-Actor-7B or GUI-AIMA-3B models, respectively.
All other components of Trifuse, including token and head filtering, CS fusion,
and two-stage localization, remain unchanged.

\begin{table}[]
	\centering
	\tabcolsep=2.5pt
	\renewcommand{\arraystretch}{1.05}
	\caption{Performance comparison on four grounding benchmarks. TF denotes training-free methods, while TB denotes training-based methods. $\Delta$ denotes the improvement of Trifuse over the corresponding baseline.}
	\label{tab:all}
	\scalebox{0.55}{
\begin{tabular}{ccccccccccc}
	\hline
	\multicolumn{2}{c}{\multirow{2}{*}{Method}} & \multirow{2}{*}{\begin{tabular}[c]{@{}c@{}}Model\\ Size\end{tabular}} & \multicolumn{2}{c}{ScreenSpot} & \multicolumn{2}{c}{ScreenSpot-v2} & \multicolumn{2}{c}{ScreenSpot-Pro} & \multicolumn{2}{c}{OSWorld-G} \\ \cline{4-11} 
	\multicolumn{2}{c}{} &  & Avg. & $\Delta$ & Avg. & $\Delta$ & Avg. & $\Delta$ & Avg. & $\Delta$ \\ \hline
	\multirow{3}{*}{TF} & TAG~\cite{tag} & 8.5B & 57.5 & - & 51.2 & - & 3.0 & - & 25.3 & - \\
	& Trifuse & 3B & 81.1 & \textbf{+23.6} & 82.6 & \textbf{+31.4} & 18.9 & \textbf{+15.9} & 38.7 & \textbf{+13.4} \\
	& Trifuse & 7B & 86.2 & \textbf{+28.7} & 86.9 & \textbf{+35.7} & 29.7 & \textbf{+26.7} & 43.6 & \textbf{+18.3} \\ \hline
	\multirow{6}{*}{TB} & GUI-Actor~\cite{guiactor} & 3B & 86.5 & - & 91.0 & - & 42.2 & - & 54.6 & - \\
	& Trifuse+GUI-Actor & 3B & 89.5 & \textbf{+3.0} & 91.5 & \textbf{+0.5} & 42.7 & \textbf{+0.5} & 56.5 & \textbf{+1.9} \\
	& GUI-Actor~\cite{guiactor} & 7B & 88.3 & - & 92.1 & - & 44.6 & - & 56.6 & - \\
	& Trifuse+GUI-Actor & 7B & 90.5 & \textbf{+2.2} & 93.2 & \textbf{+1.1} & 45.8 & \textbf{+1.2} & 57.9 & \textbf{+1.3} \\
	& GUI-AIMA~\cite{guiaima} & 3B & 88.1 & - & 91.5 & - & 49.8 & - & 58.3 & - \\
	& Trifuse+GUI-AIMA & 7B & 90.6 & \textbf{+2.5} & 92.9 & \textbf{+1.4} & 51.3 & \textbf{+1.5} & 58.4 & \textbf{+0.1} \\ \hline
\end{tabular}
}

\end{table}
\begin{table}[]
	\centering
	\tabcolsep=2.5pt
	\renewcommand{\arraystretch}{1.05}
	\caption{Performance comparison on ScreenSpot. Training Data Size refer to the number of images used for training (expect UI-TARS).}
	\scalebox{0.6}{
\begin{tabular}{ccccccccccccc}
	\hline
	\multicolumn{3}{c}{\multirow{2}{*}{Method}} & \multirow{2}{*}{\begin{tabular}[c]{@{}c@{}}Training \\ Data Size\end{tabular}} & \multirow{2}{*}{\begin{tabular}[c]{@{}c@{}}Model \\ Size\end{tabular}} & \multicolumn{2}{c}{Mobile} & \multicolumn{2}{c}{Desktop} & \multicolumn{2}{c}{Web} & \multirow{2}{*}{Avg.} & \multirow{2}{*}{$\Delta$} \\ \cline{6-11}
	\multicolumn{3}{c}{} &  &  & Text & Icon & Text & Icon & Text & Icon &  &  \\ \hline
	\multicolumn{2}{c}{\multirow{4}{*}{\rotatebox{90}{General}}} & GPT-4o & - & / & 30.5 & 23.2 & 20.6 & 19.4 & 11.1 & 7.8 & 18.8 & - \\
	\multicolumn{2}{c}{} & Qwen2.5-VL & - & 3B & 62.1 & 46.4 & 54.1 & 30.0 & 31.2 & 48.3 & 46.9 & - \\
	\multicolumn{2}{c}{} & Qwen2.5-VL & - & 7B & 93.4 & 76.4 & 87.6 & 57.9 & 82.2 & 63.1 & 78.6 & - \\
	\multicolumn{2}{c}{} & OmniParser & - & / & 93.9 & 57.0 & 91.3 & 63.6 & 81.3 & 51.0 & 73.0 & - \\ \hline
	\multicolumn{2}{c}{\multirow{8}{*}{\rotatebox{90}{SFT}}} & MP-GUI & 0.68M & 8B & 86.8 & 65.9 & 70.8 & 56.4 & 58.3 & 46.6 & 64.1 & - \\
	\multicolumn{2}{c}{} & UGround & 1.3M & 7B & 82.8 & 60.3 & 82.5 & 63.6 & 80.4 & 70.4 & 73.3 & - \\
	\multicolumn{2}{c}{} & ShowUI & 0.26M & 2B & 91.6 & 69.0 & 81.8 & 59.0 & 83.0 & 65.5 & 74.9 & - \\
	\multicolumn{2}{c}{} & Aguvius & 1M & 7B & 78.2 & {\ul 88.3} & 70.7 & {\ul 88.1} & 74.8 & 85.7 & 81.8 & - \\
	\multicolumn{2}{c}{} & UI-TARS & 50B Token & 2B & 93.0 & 75.5 & 90.7 & 68.6 & 84.3 & 74.8 & 82.3 & - \\
	\multicolumn{2}{c}{} & UI-TARS & 50B Token & 7B & 94.5 & 85.2 & 95.9 & 85.7 & 90.0 & 83.5 & 89.5 & - \\
	\multicolumn{2}{c}{} & JEDI & 1.39M & 3B & 96.9 & 81.5 & {\ul 96.9} & 78.6 & 88.5 & 83.7 & 88.6 & - \\
	\multicolumn{2}{c}{} & JEDI & 1.39M & 7B & 96.9 & 87.2 & 95.9 & 87.9 & 94.4 & 84.2 & {\ul 91.7} & - \\ \hline
	\multicolumn{2}{c}{\multirow{4}{*}{\rotatebox{90}{RL}}} & UI-R1 & 136 & 3B & 95.6 & 84.7 & 85.2 & 73.3 & 90.2 & 59.3 & 83.3 & - \\
	\multicolumn{2}{c}{} & GUI-R1 & 3K & 3B & 96.7 & 76.7 & 89.6 & 72.1 & 93.8 & 64.8 & 83.6 & - \\
	\multicolumn{2}{c}{} & GUI-G1 & 17K & 3B & \textbf{98.6} & 85.8 & 96.4 & 80.7 & 91.4 & 82.3 & 90.3 & - \\
	\multicolumn{2}{c}{} & GUI-G$^2$ & 100K & 7B & 96.7 & \textbf{90.8} & 95.9 & \textbf{88.6} & 90.9 & {\ul 86.9} & \textbf{92.0} & - \\ \hline
	\multirow{9}{*}{\rotatebox{90}{\textbf{{Attention-based}}}} & \multirow{3}{*}{TF} & TAG & 0 & 8.5B & 88.3 & 29.3 & 82.5 & 28.6 & 70.9 & 29.1 & 57.5 & - \\
	&  & Trifuse & 0 & 3B & 91.9 & 73.4 & 90.2 & 65.7 & 87.4 & 70.4 & 81.1 & \textbf{+23.6} \\
	&  & Trifuse & 0 & 7B & 92.7 & 79.0 & 90.2 & 80.0 & 91.7 & 79.6 & 86.2 & \textbf{+28.7} \\ \cline{2-13} 
	& \multirow{6}{*}{TB} & GUI-Actor & 1M & 3B & 93.0 & 79.9 & 88.1 & 78.6 & 90.9 & 84.0 & 86.5 & - \\
	&  & Trifuse+GUI-Actor & - & 3B & 96.0 & 82.5 & 90.7 & 80.7 & 96.5 & 85.4 & 89.5 & \textbf{+3.0} \\
	&  & GUI-Actor & 1M & 7B & 94.9 & 82.1 & 91.8 & 80.0 & 91.3 & 85.4 & 88.3 & - \\
	&  & Trifuse+GUI-Actor & - & 7B & 97.8 & 84.3 & 90.7 & 82.9 & \textbf{95.7} & \textbf{86.9} & 90.5 & \textbf{+2.2} \\
	&  & GUI-AIMA & 85K & 3B & 96.3 & 83.8 & 94.3 & 85.7 & 92.6 & 72.3 & 88.1 & - \\
	&  & Trifuse+GUI-AIMA & - & 3B & {\ul 98.2} & 86.0 & \textbf{96.9} & 85.0 & {\ul 94.8} & 79.1 & 90.6 & \textbf{+2.5} \\ \hline
\end{tabular}
}

	\label{tab:screenspot}
\end{table}

\subsection{Main Results}
We present the evaluation results of Trifuse against state-of-the-art GUI grounding methods across four benchmarks in Table~\ref{tab:screenspot}, Table~\ref{tab:screenspotv2}, Table~\ref{tab:screenspotpro} and Table~\ref{tab:osworldg}. 
To facilitate a clearer comparison with attention-based approaches,
Table~\ref{tab:all} further summarizes the results under matched model
backbones and evaluation settings.
We emphasize pairwise comparisons under identical model settings and training paradigms, highlighting the consistent improvements brought by Trifuse over its attention-based counterparts rather than reporting global maxima across different approaches.

Several key observations can be drawn from the results.
(1) Despite requiring no task-specific fine-tuning, Trifuse achieves competitive or superior performance compared to supervised fine-tuned (SFT) models and even outperforms several reinforcement learning (RL)-based methods across all benchmarks. 
(2) Using a 3B-parameter backbone, Trifuse attains higher average accuracy than existing training-free attention-based methods,
substantially surpassing the TAG baseline~\cite{tag}.
(3) When integrated with existing attention-based methods such as GUI-Actor
and GUI-AIMA~\cite{guiactor,guiaima}, Trifuse consistently improves grounding accuracy, demonstrating
its effectiveness as a modular enhancement.

Overall, these results validate that the proposed Consensus-SinglePeak (CS)
fusion strategy effectively captures complementary information from attention,
OCR, and caption modalities.
Trifuse delivers robust performance gains across diverse platforms,
resolutions, and element types, while remaining entirely independent of GUI-specific annotated data.
\begin{table}[]
	\centering
	\tabcolsep=2.5pt
	\renewcommand{\arraystretch}{1.05}
	\caption{Performance comparison on ScreenSpot-v2.}
	\scalebox{0.60}{
\begin{tabular}{ccccccccccccc}
	\hline
	\multicolumn{3}{c}{\multirow{2}{*}{Method}} & \multirow{2}{*}{\begin{tabular}[c]{@{}c@{}}Training \\ Data Size\end{tabular}} & \multirow{2}{*}{\begin{tabular}[c]{@{}c@{}}Model \\ Size\end{tabular}} & \multicolumn{2}{c}{Mobile} & \multicolumn{2}{c}{Desktop} & \multicolumn{2}{c}{Web} & \multirow{2}{*}{Avg.} & \multirow{2}{*}{$\Delta$} \\ \cline{6-11}
	\multicolumn{3}{c}{} &  &  & Text & Icon & Text & Icon & Text & Icon &  &  \\ \hline
	\multicolumn{2}{c}{\multirow{4}{*}{\rotatebox{90}{General}}} & Qwen2.5-VL & - & 3B & 64.8 & 49.8 & 62.4 & 32.9 & 59.6 & 50.5 & 55.0 & - \\
	\multicolumn{2}{c}{} & Qwen2.5-VL & - & 7B & 90.8 & 73.4 & 85.1 & 62.9 & 75.2 & 64.1 & 76.6 & - \\
	\multicolumn{2}{c}{} & Operator & - & - & 48.3 & 41.5 & 90.2 & 80.3 & 92.8 & 84.3 & 70.5 & - \\
	\multicolumn{2}{c}{} & OmniParser-v2 & - & - & 95.5 & 74.6 & 92.3 & 60.9 & 88.0 & 59.6 & 80.7 & - \\ \hline
	\multicolumn{2}{c}{\multirow{4}{*}{\rotatebox{90}{SFT}}} & UI-TARS & 50B Token & 2B & 95.2 & 79.1 & 90.7 & 68.6 & 87.2 & 78.3 & 84.7 & - \\
	\multicolumn{2}{c}{} & UI-TARS & 50B Token & 7B & 96.9 & 89.1 & 95.4 & 85.0 & 93.6 & 85.2 & 91.6 & - \\
	\multicolumn{2}{c}{} & JEDI & 1.39M & 3B & 96.9 & 81.5 & 96.9 & 78.6 & 88.5 & 83.7 & 88.6 & - \\
	\multicolumn{2}{c}{} & JEDI & 1.39M & 7B & 96.9 & 87.2 & 95.9 & 87.9 & 94.4 & 84.2 & 91.7 & - \\ \hline
	\multicolumn{2}{c}{\multirow{3}{*}{\rotatebox{90}{RL}}} & UI-R1 & 136 & 3B & 96.2 & 84.3 & 92.3 & 63.6 & 89.2 & 75.4 & 85.4 & - \\
	\multicolumn{2}{c}{} & GUI-R1 & 3K & 3B & 97.6 & 78.2 & 94.3 & 64.3 & 91.0 & 72.4 & 85.0 & - \\
	\multicolumn{2}{c}{} & GUI-G$^2$ & 100K & 7B & {\ul 98.3} & \textbf{91.9} & 95.4 & {\ul 89.3} & 94.0 & {\ul 87.7} & \textbf{93.3} & - \\ \hline
	\multirow{9}{*}{\rotatebox{90}{\textbf{Attention-based}}} & \multirow{3}{*}{TF} & TAG & 0 & 8.5B & 78.8 & 24.0 & 72.2 & 25.0 & 67.4 & 25.8 & 51.2 & - \\
	&  & Trifuse & 0 & 3B & 92.7 & 75.1 & 92.8 & 64.3 & 91.3 & 70.9 & 82.6 & \textbf{+31.4} \\
	&  & Trifuse & 0 & 7B & 95.2 & 79.0 & 93.3 & 79.3 & 92.2 & 78.2 & 86.9 & \textbf{+35.7} \\ \cline{2-13} 
	& \multirow{6}{*}{TB} & GUI-Actor & 1M & 2B & 97.6 & 83.4 & 96.9 & 83.6 & 94.0 & 85.7 & 91.0 & - \\
	&  & Trifuse+GUI-Actor & - & 2B & 97.8 & 84.3 & 96.4 & 85.7 & 94.8 & 86.9 & 91.5 & \textbf{+0.5} \\
	&  & GUI-Actor & 1M & 7B & 97.6 & 88.2 & 96.9 & 85.7 & 93.2 & 86.7 & 92.1 & - \\
	&  & Trifuse+GUI-Actor & - & 7B & 98.2 & {\ul 90.8} & \textbf{97.4} & 87.1 & 93.9 & \textbf{88.3} & {\ul 93.2} & \textbf{+1.1} \\
	&  & GUI-AIMA & 85K & 3B & \textbf{99.2} & 85.9 & 96.1 & 88.9 & {\ul 96.1} & 80.2 & 91.5 & - \\
	&  & Trifuse+GUI-AIMA & - & 3B & 98.2 & 87.3 & {\ul 96.9} & \textbf{90.7} & \textbf{97.0} & 85.4 & 92.9 & \textbf{+1.4} \\ \hline
\end{tabular}
	}
	
	\label{tab:screenspotv2}
\end{table}
\begin{table}[]
	\centering
	\renewcommand{\arraystretch}{1.05}
	\tabcolsep=2pt
	\caption{Performance comparison on ScreenSpot-Pro.}
	\scalebox{0.55}{
		\begin{tabular}{ccccccccccccc}
			\hline
			\multicolumn{3}{c}{Method} & \begin{tabular}[c]{@{}c@{}}Training \\ Data Size\end{tabular} & \begin{tabular}[c]{@{}c@{}}Model \\ Size\end{tabular} & CAD & Dev & Creative & Scientific & Office & OS & Avg. & $\Delta$ \\ \hline
			\multicolumn{2}{c}{\multirow{4}{*}{\rotatebox{90}{General}}} & GPT-4o & - & / & 1.5 & 0.7 & 0.6 & 1.3 & 0.8 & 0.0 & 0.8 & - \\
			\multicolumn{2}{c}{} & Qwen2.5-VL & - & 3B & 8.7 & 12.1 & 17.1 & 26.3 & 29.6 & 6.1 & 16.1 & - \\
			\multicolumn{2}{c}{} & Qwen2.5-VL & - & 7B & 13.2 & 26.1 & 24.8 & 33.1 & 45.2 & 23.5 & 26.8 & - \\
			\multicolumn{2}{c}{} & Claude & - & / & 11.9 & 13.2 & 17.1 & 26.9 & 26.9 & 8.0 & 17.1 & - \\ \hline
			\multicolumn{2}{c}{\multirow{4}{*}{\rotatebox{90}{SFT}}} & UGround-v1 & 1.3M & 7B & 12.3 & 28.1 & 32.6 & 41.0 & 49.6 & 24.5 & 31.1 & - \\
			\multicolumn{2}{c}{} & UI-TARS & 50B Token & 72B & 17.3 & 40.7 & 40.7 & 47.8 & 54.8 & 30.1 & 38.1 & - \\
			\multicolumn{2}{c}{} & JEDI & 1.39M & 3B & 23.1 & 38.1 & 35.8 & 40.3 & 57.0 & 25.0 & 36.1 & - \\
			\multicolumn{2}{c}{} & JEDI & 1.39M & 7B & 31.6 & 36.1 & 41.3 & 49.5 & 65.7 & 33.2 & 42.0 & - \\ \hline
			\multicolumn{2}{c}{\multirow{3}{*}{\rotatebox{90}{RL}}} & UI-R1 & 136 & 3B & 10.0 & 13.7 & 17.9 & 30.6 & 27.4 & 9.2 & 17.8 & - \\
			\multicolumn{2}{c}{} & GUI-G1 & 17K & 3B & 32.4 & 31.1 & 26.9 & 49.6 & 59.1 & 17.6 & 37.1 & - \\
			\multicolumn{2}{c}{} & GUI-G$^2$ & 100K & 7B & 17.7 & 22.8 & 27.4 & 30.8 & 39.1 & 17.9 & 25.5 & - \\ \hline
			\multirow{9}{*}{\rotatebox{90}{Attention-based}} & \multirow{3}{*}{TF} & TAG & 0 & 8.5B & 5.4 & 2.0 & 2.9 & 2.6 & 2.6 & 2.0 & 3.0 & - \\
			&  & Trifuse & 0 & 3B & 17.1 & 14.1 & 17.7 & 22.1 & 30.4 & 12.8 & 18.9 & \textbf{+15.9} \\
			&  & Trifuse & 0 & 7B & 28.8 & 27.7 & 22.7 & 31.2 & 47.8 & 26.5 & 29.7 & \textbf{+26.7} \\ \cline{2-13} 
			& \multirow{6}{*}{TB} & GUI-Actor & 1M & 3B & 34.1 & 39.8 & 36.7 & 49.6 & 61.3 & 35.2 & 42.2 & - \\
			&  & Trifuse+GUI-Actor & - & 3B & 35.2 & 41.1 & 37.5 & 48.6 & 63.4 & 34.7 & 42.7 & \textbf{+0.5} \\
			&  & GUI-Actor & 1M & 7B & 38.3 & 38.1 & 41.4 & 50.8 & 63.0 & 38.8 & 44.6 & - \\
			&  & Trifuse+GUI-Actor & - & 7B & {\ul 40.0} & 40.5 & 42.0 & 53.3 & 61.7 & 40.8 & 45.8 & \textbf{+1.2} \\
			&  & GUI-AIMA & 85K & 3B & 39.3 & {\ul 48.9} & {\ul 44.7} & \textbf{57.0} & {\ul 65.6} & {\ul 50.0} & {\ul 49.8} & - \\
			&  & Trifuse+GUI-AIMA & - & 3B & \textbf{42.0} & \textbf{51.2} & \textbf{45.4} & {\ul 56.2} & \textbf{67.8} & \textbf{51.5} & \textbf{51.3} & \textbf{+1.5} \\ \hline
		\end{tabular}
	}
	
	\label{tab:screenspotpro}
\end{table}

\begin{table}[]
	\centering
	\tabcolsep=3pt
	\caption{Performance comparison on OSWorld-G.}
	\scalebox{0.48}{
\begin{tabular}{ccccccccccc}
	\hline
	\multicolumn{3}{c}{Model} & \begin{tabular}[c]{@{}c@{}}Training \\ Data Size\end{tabular} & \begin{tabular}[c]{@{}c@{}}Model \\ Size\end{tabular} & \begin{tabular}[c]{@{}c@{}}Text \\ Matching\end{tabular} & \begin{tabular}[c]{@{}c@{}}Element \\ Recognition\end{tabular} & \begin{tabular}[c]{@{}c@{}}Layout \\ Understanding\end{tabular} & \begin{tabular}[c]{@{}c@{}}Fine-grained \\ Manipulation\end{tabular} & Avg. & $\Delta$ \\ \hline
	\multicolumn{2}{c}{\multirow{4}{*}{\rotatebox{90}{General}}} & Operator & - & - & 51.3 & 42.4 & 46.6 & 31.5 & 40.6 & - \\
	\multicolumn{2}{c}{} & Gemini-2.5-Pro & - & - & 59.8 & 45.5 & 49.0 & 33.6 & 45.2 & - \\
	\multicolumn{2}{c}{} & Qwen2.5-VL & - & 3B & 41.4 & 28.8 & 34.8 & 13.4 & 27.3 & - \\
	\multicolumn{2}{c}{} & Qwen2.5-VL & - & 7B & 45.6 & 32.7 & 41.9 & 18.1 & 31.4 & - \\ \hline
	\multicolumn{2}{c}{\multirow{4}{*}{\rotatebox{90}{SFT}}} & UGround-V1 & 1.3M & 7B & 51.3 & 40.3 & 43.5 & 24.8 & 36.4 & - \\
	\multicolumn{2}{c}{} & UI-TARS & 50B Token & 7B & 60.2 & 51.8 & 54.9 & 35.6 & 47.5 & - \\
	\multicolumn{2}{c}{} & JEDI & 1.39M & 3B & 67.4 & 53.0 & 53.8 & 44.3 & 50.9 & - \\
	\multicolumn{2}{c}{} & JEDI & 1.39M & 7B & 65.9 & 55.5 & 57.7 & 46.9 & 54.1 & - \\ \hline
	\multirow{9}{*}{\rotatebox{90}{\textbf{Attention-based}}} & \multirow{3}{*}{TF} & TAG & 0 & 8.5B & 44.9 & 19.2 & 28.1 & 8.4 & 25.3 & - \\
	&  & Trifuse & 0 & 3B & 55.2 & 33.0 & 45.6 & 24.7 & 38.7 & \textbf{+13.4} \\
	&  & Trifuse & 0 & 7B & 65.3 & 35.9 & 50.4 & 27.2 & 43.6 & \textbf{+18.3} \\ \cline{2-11} 
	& \multirow{6}{*}{TB} & GUI-Actor & 1M & 3B & 64.4 & 60.6 & 64.8 & 33.6 & 54.6 & - \\
	&  & Trifuse+GUI-Actor & - & 3B & 64.9 & 61.4 & 66.7 & 34.4 & 56.5 & \textbf{+1.9} \\
	&  & GUI-Actor & 1M & 7B & {\ul 65.9} & 62.7 & 66.4 & {\ul 38.2} & 56.6 & - \\
	&  & Trifuse+GUI-Actor & - & 7B & \textbf{66.4} & 62.3 & 67.1 & \textbf{38.9} & 57.9 & \textbf{+1.3} \\
	&  & GUI-AIMA & 85K & 3B & 64.8 & \textbf{65.5} & {\ul 68.8} & 36.8 & {\ul 58.3} &  \\
	&  & Trifuse+GUI-AIMA & - & 3B & 65.6 & {\ul 63.7} & \textbf{69.4} & 36.3 & \textbf{58.4} & \textbf{+0.1} \\ \hline
\end{tabular}
	}
	
	\label{tab:osworldg}
\end{table}
\subsection{Ablation Studies}
\label{label:ablation}
We perform systematic ablation studies to examine the contribution of the core components of Trifuse, including the attention extraction strategy (token and head-level filtering) and the proposed Consensus-SinglePeak (CS) fusion strategy.
These studies are designed to isolate the effect of each component and to
assess their individual and combined impact on grounding performance.
Unless otherwise specified, all ablation experiments adopt
Qwen2.5-VL-3B-Instruct as the backbone.
Additional ablation results, including sensitivity analyses of key
hyperparameters and ablation studies on two-stage localization strategy are reported in Appendix~\ref{lab:add_experiment}.

\paragraph{Attention Modality}
We analyze the effect of token filtering and head filtering on grounding
performance when using the attention modality alone.
Table~\ref{tab:attn_ablation} reports the results under different selection
strategies.

\textbf{Head filtering.}
With a fixed token selection strategy, head filtering has a pronounced impact
on grounding accuracy.
Aggregating attention from all heads leads to substantial performance
degradation, indicating that many attention heads encode information that is
irrelevant to spatial localization.
Restricting attention to a subset of heads (layer 16 to layer 32) yields moderate improvements even without explicit ranking.
In contrast, our proposed top-$k$ head selection strategy, which ranks heads by
spatial entropy, consistently achieves the best performance by retaining only
the most spatially informative heads.

\textbf{Token filtering.} 
With a fixed head selection strategy, token filtering also plays a critical role.
Aggregating attention across all instruction tokens introduces noise from task-irrelevant words, which dilutes the spatial grounding signal.
Using only the final token partially alleviates this issue but may discard important semantic cues related to the target element.
Our top-$k$ token selection strategy, which ranks tokens according to their visual relevance scores, achieves the strongest performance by focusing on semantically informative tokens.

Overall, these results demonstrate that selective attention extraction through
joint filtering of both tokens and heads is crucial for effective attention-based
GUI grounding.

\begin{table}[]
	\centering
	\tabcolsep=1.5pt
	\renewcommand{\arraystretch}{1.3}
	\caption{Ablation studies on token and head selection strategies for attention extraction.}
	\scalebox{0.65}{
		\begin{tabular}{cccccccccccccc}
			\hline
			\multicolumn{2}{c}{\multirow{2}{*}{Method}} & \multicolumn{4}{c}{ScreenSpot} & \multicolumn{4}{c}{ScreenSpot-v2} & \multicolumn{2}{c}{ScreenSpot-Pro} & \multicolumn{2}{c}{OSWorld-G} \\ \cline{3-14} 
			\multicolumn{2}{c}{} & Text & Icon & Avg. & $\Delta$ & Text & Icon & Avg. & $\Delta$ & Avg. & $\Delta$ & Avg. & $\Delta$ \\ \hline
			\multirow{3}{*}{\rotatebox{90}{Top Token}} & All Head & 13.2 & 6.6 & 10.2 & - & 11.5 & 6.2 & 9.2 & - & 3.6 & - & 7.3 & - \\
			& Range Head & 18.4 & 10.8 & 14.9 & \textbf{+4.7} & 17.0 & 9.8 & 13.8 & \textbf{+4.6} & 6.2 & \textbf{+1.6} & 14.9 & \textbf{+7.6} \\
			& Top Head & 66.5 & 48.7 & 58.5 & \textbf{+48.3} & 62.2 & 45.1 & 54.7 & \textbf{+45.5} & 8.0 & \textbf{+4.4} & 28.0 & \textbf{+20.7} \\ \hline
			\multirow{3}{*}{\rotatebox{90}{Top Head}} & All Token & 50.9 & 32.3 & 42.5 & - & 49.9 & 29.9 & 40.9 & - & 5.8 & - & 18.6 & - \\
			& Last Token & 61.6 & 44.2 & 53.7 & \textbf{+11.2} & 60.6 & 41.1 & 52.1 & \textbf{+11.2} & 7.3 & \textbf{+1.5} & 23.7 & \textbf{+5.1} \\
			& Top Token & 66.5 & 48.7 & 58.5 & \textbf{+16.0} & 62.2 & 45.1 & 54.7 & \textbf{+13.8} & 8.0 & \textbf{+2.2} & 28.0 & \textbf{+9.4} \\ \hline
		\end{tabular}
	}
	
	\label{tab:attn_ablation}
\end{table}

\paragraph{Fusion Strategy}
Table~\ref{tab:fusion_ablation_simplify} reports ablation results for different
fusion strategies.
We analyze the performance of individual modalities to reveal their complementary strengths and limitations, and then compare alternative fusion strategies.

\textbf{Single-modality analysis.} 
Each modality exhibits distinct and complementary characteristics.
The OCR modality achieves high accuracy on text-based elements but performs poorly on icon elements, reflecting its strong capability in text localization and its inherent limitation on purely visual targets.
In contrast, the caption modality provides more balanced performance by introducing semantic descriptions of visual elements; however, its overall accuracy remains limited due to imperfect caption quality and semantic
matching.
The attention modality yields the strongest single-modality performance by leveraging the MLLM’s visual ability, yet it still falls short of state-of-the-art results when used in isolation.
These observations highlight the complementary nature of the three modalities: OCR excels at textual cues, captions capture visual semantics, and attention offers broad but coarse coverage.

\textbf{Fusion mechanism comparison.} 
We compare three fusion strategies.
The Average baseline uniformly combines the three modalities, achieving modest improvements by leveraging complementary information.
However, treating all modalities equally limits its effectiveness, as the relative importance of each modality varies across task types.
The \textit{Custom} baseline applies a fixed, manually specified weighting
scheme to fuse modality-specific heatmaps, assigning weights of 0.6 to the
attention modality and 0.2 to both the OCR and caption modalities across all
benchmarks.
Despite using the same weights in all settings, this baseline exhibits
inconsistent performance, indicating that fixed fusion weights are insufficient
to adapt to the diverse characteristics of GUI grounding tasks.

In contrast, CS fusion strategy achieves substantial and consistent improvements across all benchmarks. 
Its effectiveness arises from two complementary mechanisms:
(1) the consensus term $\mathbf{M}^{\text{cons}}$ amplifies regions where multiple modalities agree, providing robustness against modality-specific noise, and (2) the single-peak term $\mathbf{M}^{\text{single}}$ preserves strong discriminative signals from individual modalities, enabling adaptation to task-specific characteristics. 
This adaptive weighting enables CS fusion to substantially outperform both single modalities and alternative fusion strategies, validating its effectiveness for GUI grounding.

\begin{table}[]
	\centering
	\tabcolsep=1.5pt
	\renewcommand{\arraystretch}{1.3}
	\caption{Ablation studies on fusion strategies. We compare single modalities (attention, OCR, caption), simple averaging, weighted fusion, and our proposed cross-modal consistency (CS) fusion.}
	\scalebox{0.65}{
		\begin{tabular}{cccccccccccccc}
			\hline
			\multicolumn{2}{c}{\multirow{2}{*}{Method}} & \multicolumn{4}{c}{ScreenSpot} & \multicolumn{4}{c}{ScreenSpot-v2} & \multicolumn{2}{c}{ScreenSpot-Pro} & \multicolumn{2}{c}{OSWorld-G} \\ \cline{3-14} 
			\multicolumn{2}{c}{} & Text & Icon & Avg. & $\Delta$ & Text & Icon & Avg. & $\Delta$ & Avg. & $\Delta$ & Avg. & $\Delta$ \\ \hline
			\multirow{3}{*}{\rotatebox{90}{Single}} & OCR & 36.3 & 5.0 & 22.1 & - & 31.9 & 3.2 & 19.7 & - & 4.9 & - & 14.3 & - \\
			& Caption & 29.8 & 22.9 & 26.7 & - & 29.0 & 16.2 & 23.2 & - & 3.7 & - & 10.2 & - \\
			& Attention & 66.5 & 48.7 & 58.5 & - & 62.2 & 45.1 & 54.7 & - & 8.0 & - & 28.0 & - \\ \hline
			\multirow{3}{*}{\rotatebox{90}{Fusion}} & Average & 78.5 & 43.4 & 63.3 & - & 83.3 & 44.5 & 65.8 & - & 13.5 & - & 33.6 &  \\
			& Custom & 63.2 & 51.2 & 58.0 & \textbf{-5.5} & 78.7 & 42.1 & 62.2 & \textbf{-3.6} & 10.7 & \textbf{-2.8} & 29.1 & -4.5 \\
			& CS (Trifuse) & 89.9 & 70.5 & 81.1 & \textbf{+17.8} & 92.3 & 71.0 & 82.6 & \textbf{+16.8} & 18.9 & \textbf{+5.4} & 43.6 & +10.0 \\ \hline
		\end{tabular}
	}
	
	\label{tab:fusion_ablation_simplify}
\end{table}


\section{Conclusion}
We introduce Trifuse, an attention-based framework for GUI grounding that
leverages complementary spatial cues through principled fusion without task-specific fine-tuning.
By systematically fusing attention, OCR-derived textual cues, and icon-level caption semantics  through the proposed Consensus-SinglePeak (CS) mechanism,
Trifuse substantially improves grounding performance over existing training-free
approaches, while approaching the performance of supervised fine-tuned methods and reinforcement learning-based methods without relying on any GUI-specific training data.
Extensive ablation studies validate the effectiveness of Trifuse’s core design choices, including (1) token filtering based on visual relevance scores to identify semantically informative query terms, (2) head filtering via spatial entropy to retain spatially focused attention heads, and (3) CS fusion, which
adaptively balances cross-modal consensus with modality-specific discriminative peaks.
Together with the proposed two-stage localization strategy, Trifuse achieves strong and robust grounding performance.
Overall, our results demonstrate that principled multimodal fusion at inference time can enable strong GUI grounding capabilities from MLLMs, highlighting a promising direction for scalable, data-efficient GUI agents.
\section*{Acknowledgments}

\section*{Impact Statement}
This paper presents work whose goal is to advance the field of machine learning by improving the reliability and generalization of graphical user interface (GUI) grounding for autonomous GUI agents. Improved GUI grounding may benefit applications such as assistive technologies, automated software testing, and human–computer interaction by enabling more accurate interpretation of diverse interfaces. 

The methods studied in this work could potentially be misused to automate interactions with software systems in unintended ways if deployed without appropriate safeguards, a risk common to GUI automation techniques in general. We believe the primary contribution of this work is methodological, providing a foundational improvement to GUI perception rather than enabling applications with direct societal risks.

\bibliography{example_paper}
\bibliographystyle{icml2026}

\newpage
\appendix
\onecolumn

\section{Benchmark Details}
\label{label:benchmarks}
\textbf{ScreenSpot}~\cite{seeclick} is a widely-used zero-shot GUI grounding benchmark spanning desktop, mobile, and web interfaces. It provides separate evaluations for text-based and icon-based (widget-level) localization tasks, enabling fine-grained analysis of grounding capabilities across different element types.

\textbf{ScreenSpot-v2}~\cite{seeclick} addresses critical quality issues in the original ScreenSpot dataset, provides cleaned annotations with verified element references and unambiguous instructions, enabling more reliable evaluation.

\textbf{ScreenSpot-Pro}~\cite{screenspotpro} extends the evaluation to high-resolution scenarios with more complex interface layouts and fine-grained element localization challenges, testing model robustness on realistic production-scale GUIs.

\textbf{OSWorld-G}~\cite{osworldg} comprises 564 meticulously annotated samples designed to systematically evaluate diverse grounding capabilities: text matching, element recognition, layout understanding, fine-grained manipulation, and infeasibility detection (identifying when target elements are absent). Each sample includes element-type annotations, providing diagnostic insights into model performance across different GUI component categories.

\textbf{ScreenSpot}, \textbf{ScreenSpot-v2}, and \textbf{ScreenSpot-Pro} categorize tasks into two types:
text and icon.
\textbf{OSWorld-G} further categorizes tasks into five types:
Text Matching, Element Recognition, Layout Understanding, Fine-grained
Manipulation, and Refusal.
The Refusal category corresponds to cases where the target element specified
by the instruction does not appear in the screenshot, and the model is
required to output a null prediction (i.e., $(-1, -1)$). This categorization also explains why Trifuse does not achieve substantial
improvements on OSWorld-G, as our method is designed for grounding visible GUI
elements and does not explicitly handle the Refusal task category.

\section{Hyperparameters Setting}
For token filtering, we set the token-visibility threshold to $\tau_v = 0.5$.

For the Single-Peak component in CS fusion, all modality-specific heatmaps are
first normalized to $[0,1]$.
A peak candidate at position $j$ for modality $s \in \{\text{attn}, \text{ocr}, \text{cap}\}$
is selected if its response exceeds a modality-wise threshold $\tau_s$.
We adopt quantile-based thresholds to ensure robustness across different
interfaces, with
$q_{\text{attn}} = 0.80$, $q_{\text{ocr}} = 0.90$, and $q_{\text{cap}} = 0.75$,
and apply a lower bound of $0.35$ to avoid excessive peak selection in flat
heatmaps.

In the confidence estimation, we set the numerical stability constant to
$\varepsilon = 10^{-6}$.
The sensitivity parameters in the sigmoid function are fixed to
$\alpha = 10$ and $\beta = 2$, which encourage high confidence when a
modality-specific peak receives non-negligible support from the remaining
modalities.
The peak amplification strength is set to $\lambda = 0.5$, resulting in a
moderate reweighting range that emphasizes reliable peaks while softly
suppressing isolated noisy responses.
Unless otherwise specified, these hyperparameters are fixed across all
experiments.

\section{Additional Experiment}
\label{lab:add_experiment}
\begin{table}[]
	\centering
	\tabcolsep=2pt
		\caption{Ablation studies on two-stage localization for final grounding method.}
	\renewcommand{\arraystretch}{1.05}
	\scalebox{1.0}{
		\begin{tabular}{ccccccccccccc}
			\hline
			\multirow{2}{*}{Method} & \multicolumn{4}{c}{ScreenSpot} & \multicolumn{4}{c}{ScreenSpot-v2} & \multicolumn{2}{c}{ScreenSpot-Pro} & \multicolumn{2}{c}{OSWorld-G} \\ \cline{2-13} 
			& Text & Icon & Avg. & $\Delta$ & Text & Icon & Avg. & $\Delta$ & Avg. & $\Delta$ & Avg. & $\Delta$ \\ \hline
			Direct & 79.8 & 64.1 & 72.7 & - & 81.8 & 67.8 & 75.7 & - & 9.2 & - & 33.1 & - \\
			Two-Stage (Trifuse) & 89.9 & 70.5 & 81.1 & \textbf{+7.4} & 92.3 & 71.0 & 82.6 & \textbf{+6.9} & 18.9 & \textbf{+9.7} & 38.7 & \textbf{+5.6} \\ \hline
		\end{tabular}
	}

	\label{tab:grounding_method}
\end{table}
Table~\ref{tab:grounding_method} compares our two-stage localization strategy against direct coordinate prediction from the fused heatmap. Trifuse consistently outperforms the direct baseline across all benchmarks.
The two-stage approach—first identifying a coarse region, then refining within a cropped window—effectively increases spatial resolution for precise localization, particularly benefiting small GUI elements on ScreenSpot-Pro where direct prediction from downsampled images struggles.
Notably, even the direct localization variant of Trifuse surpasses the TAG~\cite{tag}
baseline on all benchmarks, further validating the effectiveness of the
overall Trifuse framework.

Table~\ref{tab:ablation_screenspot}, Table~\ref{tab:fusion_ablation} and Table~\ref{tab:final_grounding} presents detailed ablation results on ScreenSpot and ScreenSpot-v2, breaking down performance by element type (text vs. icon) across platforms using Qwen2.5-VL-3B-Instruct as the base model. Performance improvements are observed across both element types, further confirming the effectiveness of fusion strategy, fine-grained token and head selection.

\begin{figure}
	\centering
	\subfloat[Head-level top-$k$ ablation]{
		\includegraphics[width=0.45\linewidth]{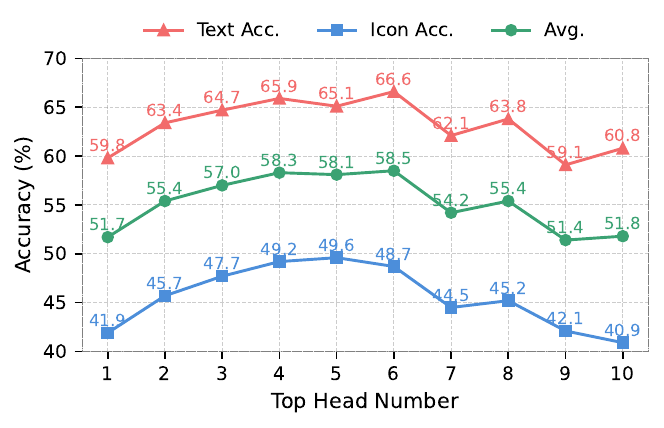}
		\label{fig:head_ablation}
	}
	\hfill
	\subfloat[Token-level top-$k$ ablation]{
		\includegraphics[width=0.45\linewidth]{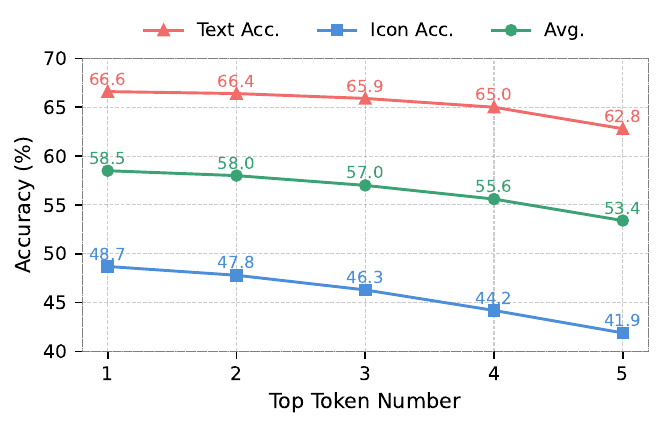}
		\label{fig:token_ablation}
	}
	\caption{Ablation studies of top-$k$ selection strategies on attention modality.}
	\label{fig:topk_ablation}
\end{figure}

Figure~\ref{fig:topk_ablation} analyzes the impact of key hyperparameters on Trifuse performance.
To isolate the effect of these parameters, we evaluate only the attention modality without fusion.
For top token selection, setting k=1 yields optimal results. Increasing k introduces irrelevant tokens that add noise and degrade grounding performance. For top head selection, performance exhibits a non-monotonic relationship: too few heads fail to capture sufficient localization signals, while too many introduce heads weakly correlated with grounding, reducing effectiveness. 
Empirically, a moderate top-$k$ head number achieves the best balance between signal coverage and noise reduction.
Based on this analysis, we set $k=1$ for tokens and $k=6$ for heads in all Trifuse experiments.
\begin{table}[t]
	\centering
	\tabcolsep=1.5pt
	\renewcommand{\arraystretch}{1.05}
			\caption{Detailed ablation studies on token and head selection strategies across ScreenSpot and ScreenSpot-v2.}
	\begin{subtable}[t]{0.48\textwidth}
		\centering
		\caption{ScreenSpot}
		\scalebox{0.9}{
		\begin{tabular}{lcccccccc}
			\hline
			\multicolumn{2}{c}{\multirow{2}{*}{Method}} 
			& \multicolumn{2}{c}{Mobile} 
			& \multicolumn{2}{c}{Desktop} 
			& \multicolumn{2}{c}{Web} 
			& \multirow{2}{*}{Avg.} \\ \cline{3-8}
			\multicolumn{2}{c}{} 
			& Text & Icon & Text & Icon & Text & Icon & \\ \hline
			\multirow{3}{*}{Top Token} & All Head & 18.7 & 7.4 & 17.0 & 12.1 & 3.5 & 1.9 & 10.2 \\
			& Range Head & 23.8 & 14.0 & 23.7 & 18.6 & 7.4 & 1.9 & 14.9 \\
			& Top Head & 79.8 & 60.2 & 62.9 & 44.3 & 53.9 & 38.8 & 58.5 \\ \hline
			\multirow{3}{*}{Top Head} & All Token & 61.9 & 39.7 & 49.5 & 39.3 & 39.1 & 19.4 & 42.5 \\
			& Last Token & 73.3 & \textbf{62.0} & 59.3 & 37.1 & 49.6 & 29.1 & 53.7 \\
			& Top Token & \textbf{79.8} & 60.2 & \textbf{62.9} & \textbf{44.3} & \textbf{53.9} & \textbf{38.8} & \textbf{58.5} \\ \hline
		\end{tabular}
		}
	\end{subtable}
	\hspace{12pt}
	\begin{subtable}[t]{0.48\textwidth}
		\centering
		\caption{ScreenSpot-v2}

		\scalebox{0.9}{
		\begin{tabular}{lcccccccc}
			\hline
			\multicolumn{2}{c}{\multirow{2}{*}{Method}} 
			& \multicolumn{2}{c}{Mobile} 
			& \multicolumn{2}{c}{Desktop} 
			& \multicolumn{2}{c}{Web} 
			& \multirow{2}{*}{Avg.} \\ \cline{3-8}
			\multicolumn{2}{c}{} 
			& Text & Icon & Text & Icon & Text & Icon & \\ \hline
			\multirow{3}{*}{Top Token} & All Head & 16.2 & 7.6 & 16.0 & 10.7 & 2.1 & 1.5 & 9.2 \\
			& Range Head & 22.8 & 12.3 & 22.2 & 16.4 & 5.6 & 2.5 & 13.8 \\
			& Top Head & 75.2 & 59.7 & 57.2 & 41.4 & 50.9 & 31.5 & 54.7 \\ \hline
			\multirow{3}{*}{Top Head} & All Token & 64.5 & 38.4 & 40.7 & 30.0 & 40.4 & 20.4 & 40.9 \\
			& Last Token & 73.4 & 54.5 & 56.7 & \textbf{42.9} & 48.7 & 25.1 & 52.1 \\
			& Top Token & \textbf{75.2} & \textbf{59.7} & \textbf{57.2} & 41.4 & \textbf{50.9} & \textbf{31.5} & \textbf{54.7} \\ \hline
		\end{tabular}
		}
	\end{subtable}

	\label{tab:ablation_screenspot}
\end{table}

\begin{table}[t]
	\centering
	\tabcolsep=1pt
	\renewcommand{\arraystretch}{1.05}
		
	\caption{Detailed ablation studies on fusion strategies across ScreenSpot and ScreenSpot-v2.}
	\begin{subtable}[t]{0.48\textwidth}
		\centering
		\caption{ScreenSpot}
		\scalebox{0.9}{
		\begin{tabular}{lcccccccc}
			\hline
			\multicolumn{2}{c}{\multirow{2}{*}{Method}} 
			& \multicolumn{2}{c}{Mobile} 
			& \multicolumn{2}{c}{Desktop} 
			& \multicolumn{2}{c}{Web} 
			& \multirow{2}{*}{Avg.} \\ \cline{3-8}
			\multicolumn{2}{c}{} 
			& Text & Icon & Text & Icon & Text & Icon & \\ \hline
			
			\multirow{3}{*}{\begin{tabular}[c]{@{}l@{}}Single\\ Modal\end{tabular}} 
			& OCR & 53.1 & 5.2 & 19.1 & 2.1 & 30.9 & 6.8 & 22.1 \\
			& Caption & 33.7 & 32.7 & 17.0 & 11.4 & 36.1 & 19.9 & 26.7 \\
			& Attention & 79.8 & 60.2 & 62.9 & 44.3 & 53.9 & 38.8 & 58.5 \\ \hline
			
			\multirow{3}{*}{Fusion} 
			& Average Fusion & 88.6 & 47.4 & 72.2 & 40.7 & 71.8 & 40.9 & 63.3 \\
			& Custom Fusion & 76.9 & 61.1 & 58.2 & 45.0 & 51.3 & 44.3 & 58.0 \\
			& CS Fusion (Trifuse) 
			& \textbf{91.9} & \textbf{73.4} & \textbf{90.2} & \textbf{65.7} & \textbf{87.4} & \textbf{70.4} & \textbf{81.1} \\ \hline
		\end{tabular}
		}
	\end{subtable}
	\hspace{12pt}
	\begin{subtable}[t]{0.48\textwidth}
		\centering
		\caption{ScreenSpot-v2}
		\scalebox{0.9}{
		\begin{tabular}{lcccccccc}
			\hline
			\multicolumn{2}{c}{\multirow{2}{*}{Method}} 
			& \multicolumn{2}{c}{Mobile} 
			& \multicolumn{2}{c}{Desktop} 
			& \multicolumn{2}{c}{Web} 
			& \multirow{2}{*}{Avg.} \\ \cline{3-8}
			\multicolumn{2}{c}{} 
			& Text & Icon & Text & Icon & Text & Icon & \\ \hline
			
			\multirow{3}{*}{\begin{tabular}[c]{@{}l@{}}Single\\ Modal\end{tabular}} 
			& OCR & 47.6 & 3.3 & 15.4 & 2.9 & 27.3 & 3.4 & 19.7 \\
			& Caption & 35.9 & 21.8 & 16.5 & 11.4 & 31.3 & 13.1 & 23.2 \\
			& Attention & 75.2 & 59.7 & 57.2 & 41.4 & 50.9 & 31.5 & 54.7 \\ \hline
			
			\multirow{3}{*}{Fusion} 
			& Average Fusion & 90.8 & 45.8 & 81.4 & 37.9 & 76.1 & 47.6 & 65.8 \\
			& Custom Fusion & 90.8 & 45.4 & 73.2 & 35.7 & 69.1 & 42.7 & 62.2 \\
			& CS Fusion (Trifuse) 
			& \textbf{92.7} & \textbf{75.1} & \textbf{92.8} & \textbf{64.3} & \textbf{91.3} & \textbf{70.9} & \textbf{82.6} \\ \hline
		\end{tabular}
		}
	\end{subtable}

	\label{tab:fusion_ablation}
\end{table}

\begin{table}[t]
	\centering
	\tabcolsep=2pt
	\renewcommand{\arraystretch}{1.05}
	
\caption{Detailed ablation studies on ScreenSpot and ScreenSpot-v2, analyzing the effect of the final grounding strategy across versions.}
		
		\begin{subtable}[t]{0.46\textwidth}
			\centering
			\caption{ScreenSpot}
			\begin{tabular}{cccccccc}
				\hline
				\multirow{2}{*}{Method} 
				& \multicolumn{2}{c}{Mobile} 
				& \multicolumn{2}{c}{Desktop} 
				& \multicolumn{2}{c}{Web} 
				& \multirow{2}{*}{Avg.} \\ \cline{2-7}
				& Text & Icon & Text & Icon & Text & Icon & \\ \hline
				Direct Location & 80.6 & 70.7 & 84.0 & 60.7 & 75.2 & 59.2 & 72.7 \\
				Two-Stage (Trifuse) 
				& \textbf{91.9} & \textbf{73.4} & \textbf{90.2} & \textbf{65.7} & \textbf{87.4} & \textbf{70.4} & \textbf{81.1} \\ \hline
			\end{tabular}
		\end{subtable}
		\hspace{25pt}
		\begin{subtable}[t]{0.46\textwidth}
			\centering
			\caption{ScreenSpot-v2}
			\begin{tabular}{cccccccc}
				\hline
				\multirow{2}{*}{Method} 
				& \multicolumn{2}{c}{Mobile} 
				& \multicolumn{2}{c}{Desktop} 
				& \multicolumn{2}{c}{Web} 
				& \multirow{2}{*}{Avg.} \\ \cline{2-7}
				& Text & Icon & Text & Icon & Text & Icon & \\ \hline
				Direct Location & 82.4 & 71.6 & 82.5 & 65.7 & 80.4 & 65.0 & 75.5 \\
				Two-Stage (Trifuse) 
				& \textbf{92.7} & \textbf{75.1} & \textbf{92.8} & \textbf{64.3} & \textbf{91.3} & \textbf{70.9} & \textbf{82.6} \\ \hline
			\end{tabular}
		\end{subtable}

	\label{tab:final_grounding}
\end{table}

\section{Limitations}
While Trifuse demonstrates strong performance for GUI grounding, several limitations suggest directions for future work.

\paragraph{Caption Model Quality}
Trifuse's overall performance depends on the quality of the OCR and caption detection models. While current OCR engines (e.g., PaddleOCR v4~\cite{paddleocr}) achieve high accuracy on text detection, existing icon caption models show limited performance. 
The lack of lightweight, high-quality GUI icon detection models constrains Trifuse's potential. 
Future work could explore improved icon captioning models or alternative visual modalities to address this bottleneck.

\paragraph{Sensitive to Hyperparameters} The Trifuse framework is sensitive to hyperparameters. 
It contains a large number of hyperparameters, such as those related to the OCR model, the icon detection model, attention modality extraction, three-modal fusion, and the final two-stage localization.
These parameters make the Trifuse framework highly sensitive to parameter variations. Our future work will focus on reducing the number of hyperparameters while maintaining the framework's localization performance.


\paragraph{Performance Ceiling}
While Trifuse substantially outperforms prior training-free methods, its performance is fundamentally constrained by the capabilities of the underlying pretrained models. Scaling to larger or more specialized backbones yields diminishing returns, suggesting inherent limitations in training-free approaches compared to GUI-specific fine-tuning. Hybrid methods that combine Trifuse's multimodal fusion with lightweight task-specific adaptation represent a promising direction for future work.

\section{Prompt}
\begin{figure}
	\centering
	\includegraphics[width=0.8\linewidth]{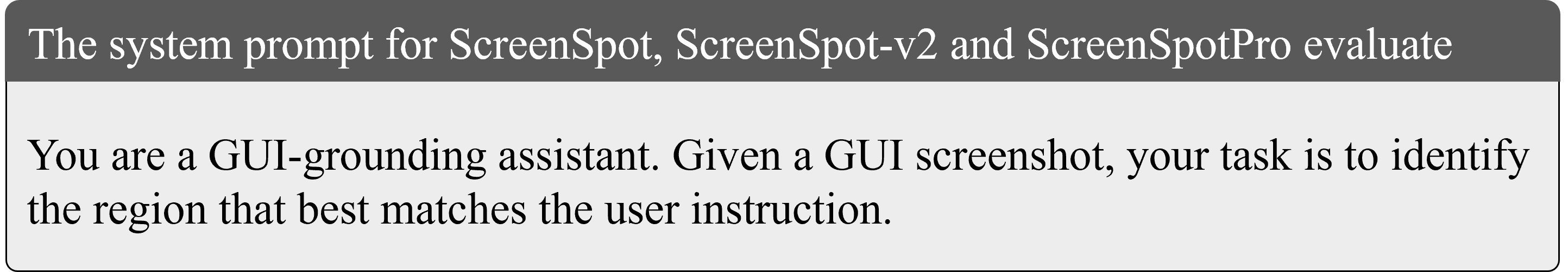}
	\caption{System prompt used for ScreenSpot, ScreenSpot-v2, and ScreenSpot-Pro evaluation.
	}
	\label{fig:prompt1}
\end{figure}

For ScreenSpot, ScreenSpot-v2 and ScreenSpot-Pro, Trifuse, GUI-Actor, and GUI-AIMA all use models from the Qwen2.5-VL series, requiring only configuration of the system prompt. Figure~\ref{fig:prompt1} shows the prompt template used across all methods.
\begin{figure}
	\centering
	\includegraphics[width=0.8\linewidth]{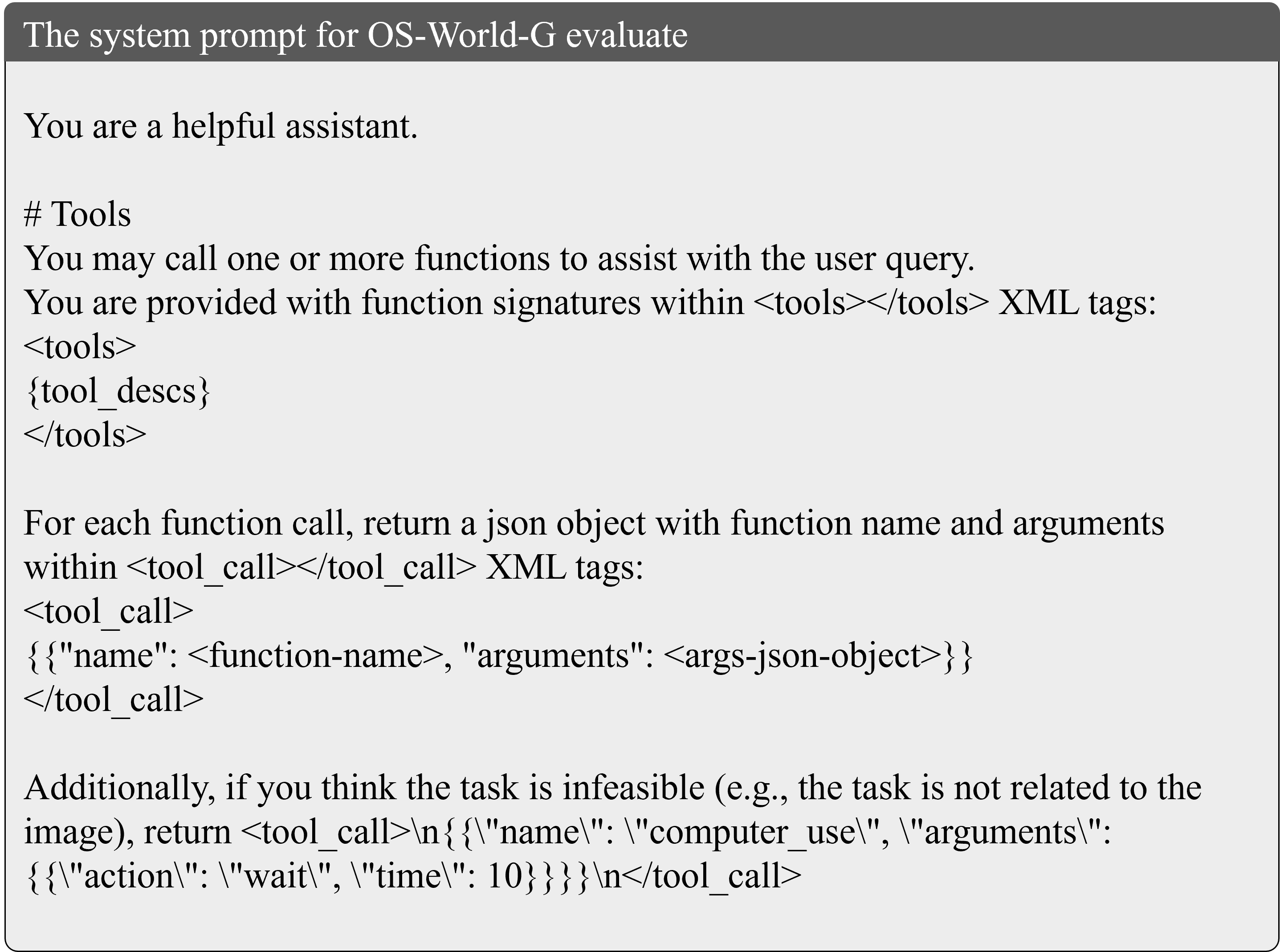}
	\caption{System prompt used for OSWorld-G evaluation.
	}
	\label{fig:prompt2}
\end{figure}
For OSWorld-G, we adopt the prompt from the original benchmark in Figure~\ref{fig:prompt2}~\cite{osworldg}. Unlike other benchmarks, OSWorld-G allows models to return (-1, -1) coordinates when no target element is found. This abstention mechanism partly explains why Trifuse shows smaller improvements on OSWorld-G compared to other benchmarks, as Trifuse does not support this output format and must always produce coordinate predictions from fused heatmap.

\section{Case Study}
\subsection{CS Fusion}
To illustrate the CS fusion strategy, we present representative examples from ScreenSpot using Qwen2.5-VL-3B-Instruct with the same modality acquisition and modality fusion described in Section~\ref{sec:method}.

\begin{figure}
	\centering
	\includegraphics[width=1\linewidth]{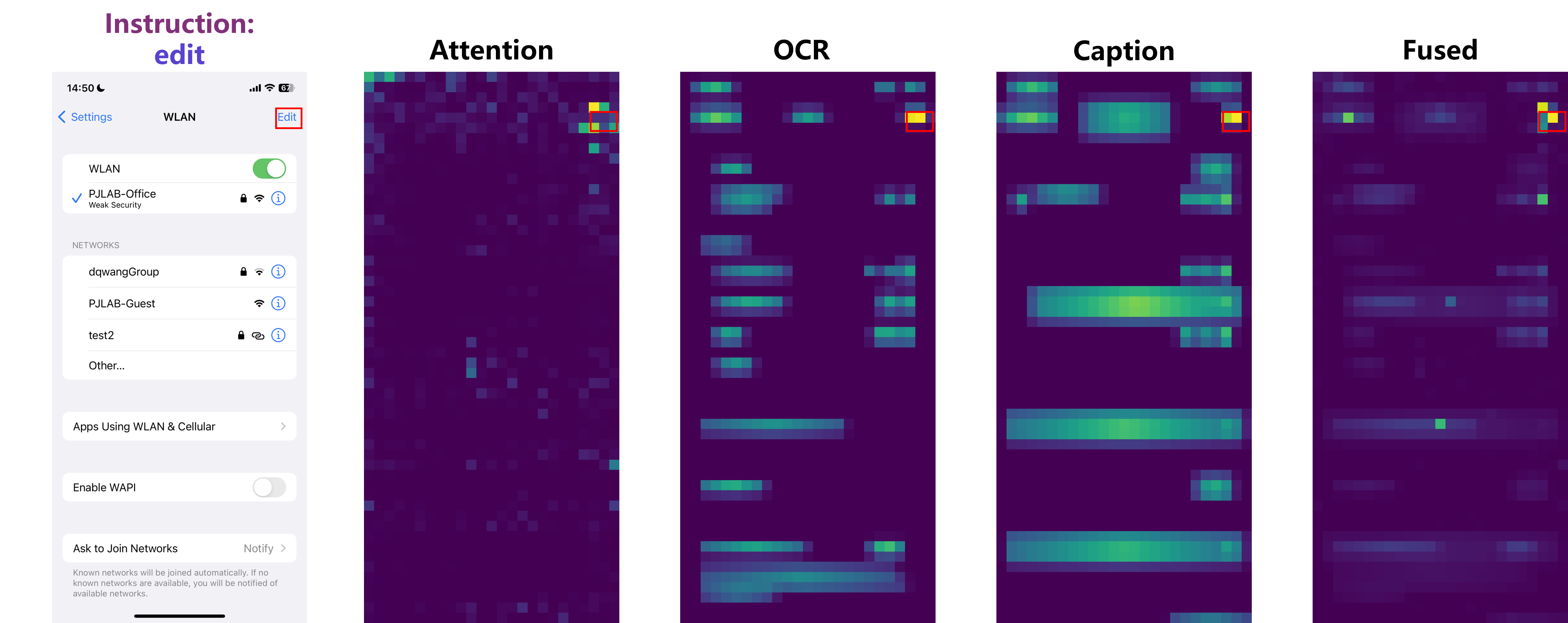}
	\caption{Visualization of CS fusion. All three modalities correctly localize the target, and fusion reinforces their agreement.
	}
	\label{fig:case1}
\end{figure}

\textbf{Consensus}. Figure~\ref{fig:case1} shows a text-based grounding task on  mobile platform. All three modalities—attention, OCR, and caption—successfully localize the target region. CS fusion produces a refined heatmap that accurately integrates these consistent signals, resulting in precise localization.

\begin{figure}
	\centering
	\includegraphics[width=1\linewidth]{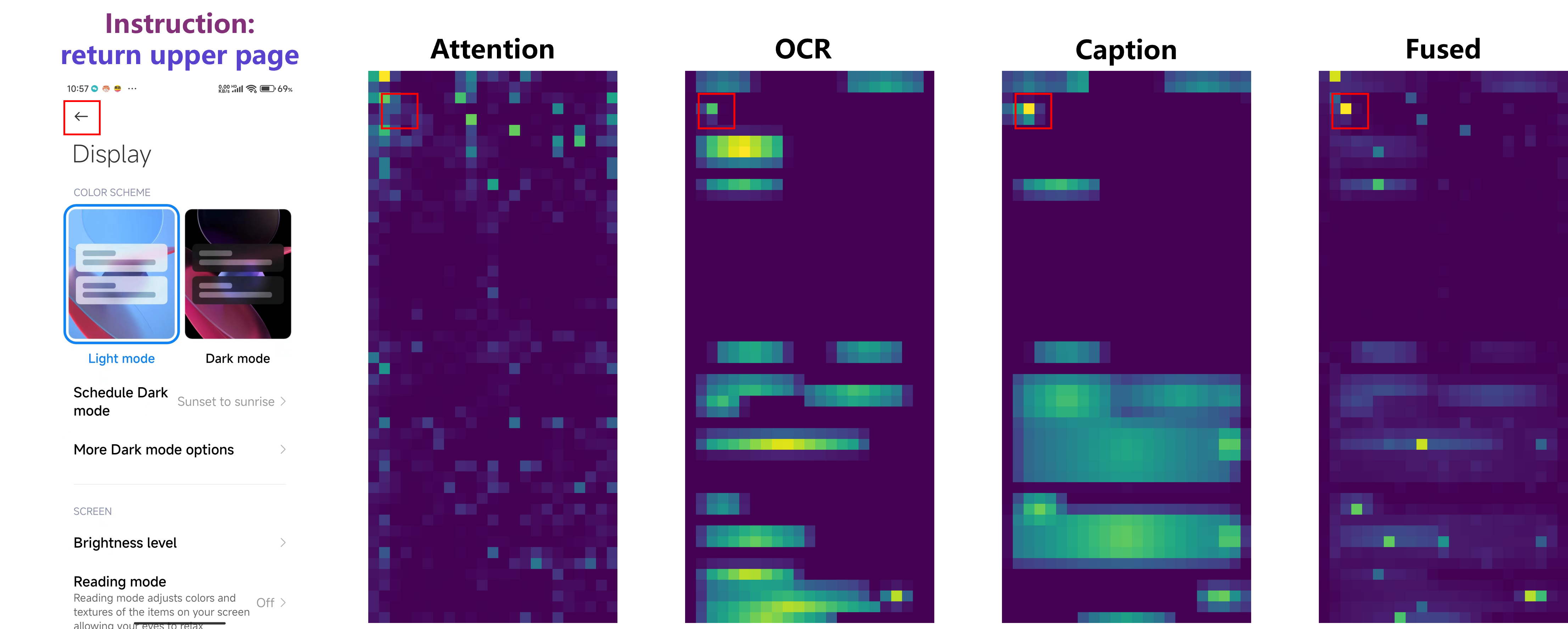}
	\caption{Visualization of CS fusion. Only caption modality correctly localizes the target; CS fusion selectively trusts the confident prediction.
	}
	\label{fig:case2}
\end{figure}

\textbf{Single-peak}. When modalities disagree, CS fusion selectively amplifies the most confident prediction. 
In Figure~\ref{fig:case2} (icon task, mobile), only the caption modality localizes the target correctly while attention and OCR fail. CS fusion identifies this single strong signal and propagates it to the final heatmap. 
Similarly, in Figure~\ref{fig:case2} (icon task, desktop), only the attention modality succeeds. CS fusion again selectively incorporates the single-peak signal, demonstrating robustness when individual modalities provide complementary but sparse information.

These cases illustrate two key properties of CS fusion: (1) when modalities agree, it reinforces consensus regions; (2) when they disagree, it selectively trusts the most confident prediction rather than averaging conflicting signals.

\begin{figure}
	\centering
	\includegraphics[width=0.8\linewidth]{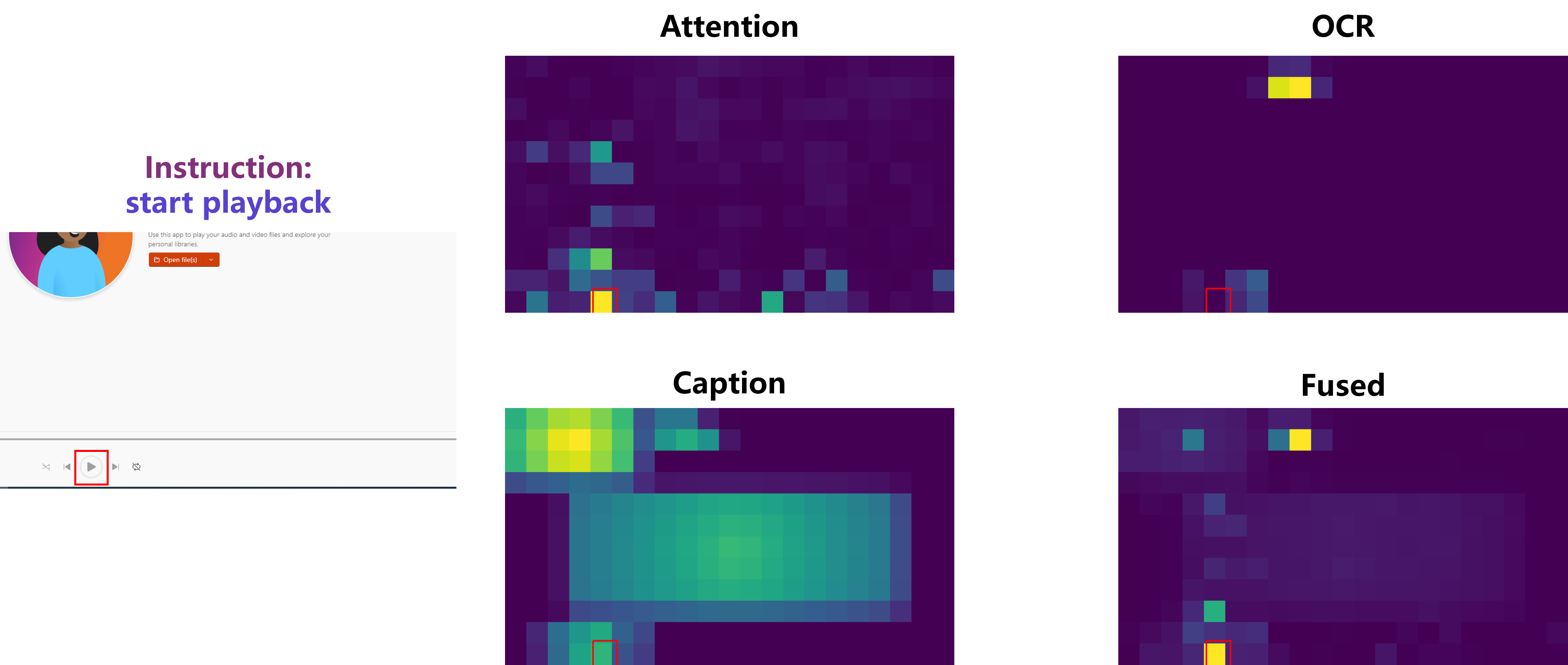}
	\caption{Visualization of CS fusion. Only attention modality correctly localizes the target; CS fusion selectively trusts the confident prediction.
	}
	\label{fig:case3}
\end{figure}

\subsection{Head filtering on Attention modality}
\label{lab:attention}
\begin{figure}
	\centering
	\includegraphics[width=0.6\linewidth]{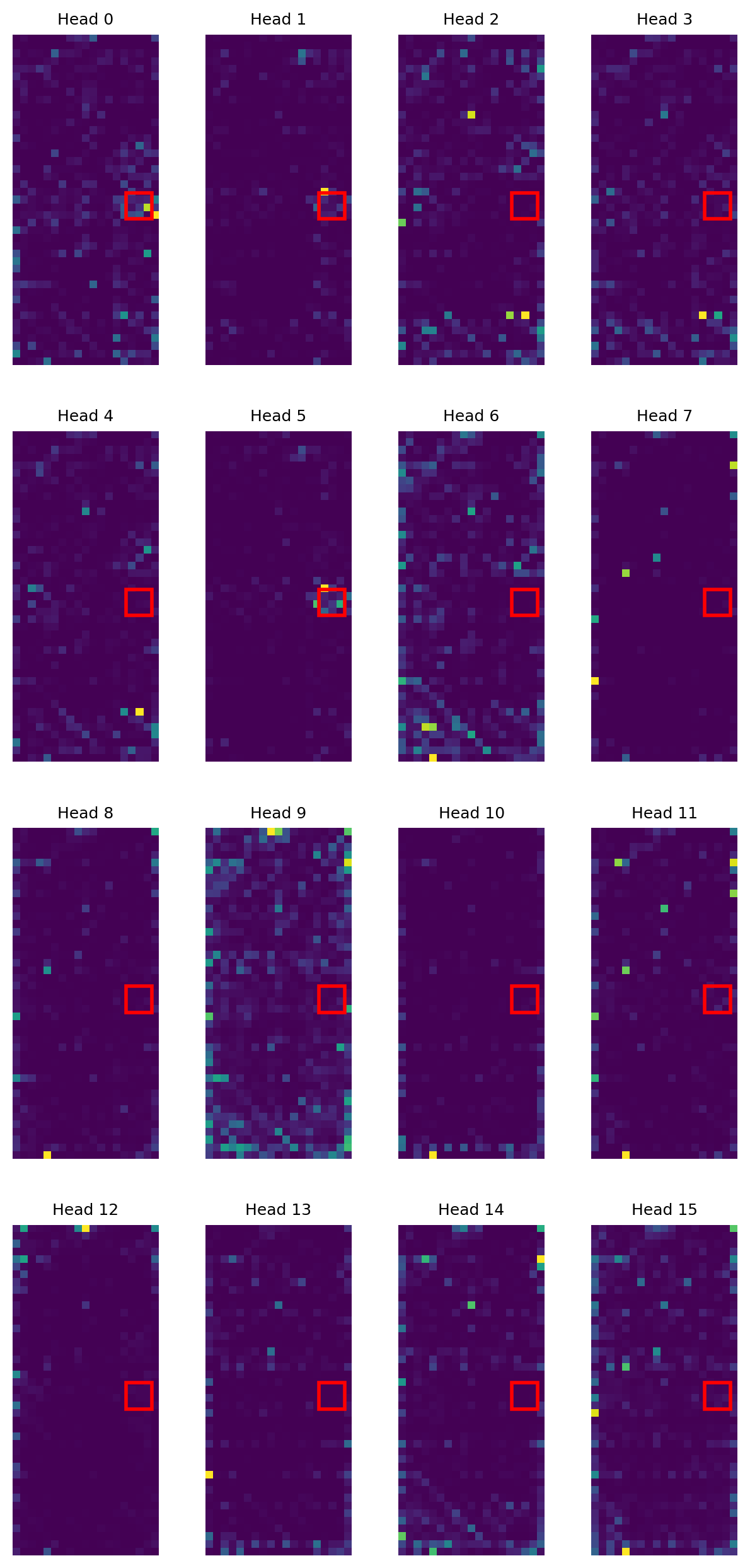}
	\caption{Visualization of attention maps across all 16 heads at layer 23 for a ScreenSpot example. The red bounding box marks the ground truth target element. 
	}
	\label{fig:heads2}
\end{figure}

To motivate the necessity of head filtering, Figure~\ref{fig:heads2} visualizes the attention maps of all 16 heads in layer 23 on a ScreenSpot example. 
The visualization reveals substantial heterogeneity across attention heads: only a small subset exhibit strong spatial alignment with the target element, while the majority of heads attend to irrelevant regions or produce diffuse attention patterns. This phenomenon is consistent across layers, where heads exhibit varying degrees of localization capability. These observations empirically motivate our head selection strategy—naive averaging across all heads would introduce significant noise from non-informative attention patterns, degrading grounding performance. By selectively aggregating attention only from heads with strong grounding signals, our approach effectively filters out this noise and enhances grounding performance.

\end{document}